\documentclass[12pt, draftclsnofoot, onecolumn]{IEEEtran}
\IEEEoverridecommandlockouts
\usepackage{amsfonts}
\usepackage{nicefrac} 
\usepackage{amsmath}
\usepackage[cmintegrals]{newtxmath}
\usepackage{urwchancal}
\usepackage{algorithm}
\usepackage{algorithmicx}
\usepackage{algpseudocode}
\usepackage{dsfont}
\usepackage{authblk}
\usepackage{bbm}
\usepackage{graphicx}
\usepackage{epstopdf}
\usepackage{subfigure}
\usepackage{stfloats}
\usepackage{eqnarray}
\usepackage{makecell}
\usepackage{multirow}
\usepackage{enumerate}
\usepackage{booktabs}
\usepackage{cite}
\usepackage{balance}
\usepackage{color}
\usepackage{bm}
\usepackage{caption}
\usepackage{mathtools}
\usepackage{url}

\DeclarePairedDelimiter\autobracket{(}{)}
\DeclarePairedDelimiter\autosquare{[}{]}
\renewcommand{\IEEEQED}{\IEEEQEDopen}
\algblock{ParFor}{EndParFor}
\algnewcommand\algorithmicparfor{\textbf{parfor}}
\algnewcommand\algorithmicpardo{\textbf{do}}
\algnewcommand\algorithmicendparfor{\textbf{end\ parfor}}
\algrenewtext{ParFor}[1]{\algorithmicparfor\ #1\ \algorithmicpardo}
\algrenewtext{EndParFor}{\algorithmicendparfor}

\newcommand{\br}[1]{\autobracket*{#1}}
\newcommand{\sq}[1]{\autosquare*{#1}}
\newcommand{\cb}[1]{\left\{#1\right\}}
\newcommand{\norm}[1]{\left\|#1\right\|}

\renewcommand{\vec}[1]{\mbox{vec}(#1)}
\newcommand{\inp}[1]{\left\langle#1\right\rangle}

\newcommand{\OO}[1]{{\cal O}\br{#1}}

\newcommand{\st}{\mbox{s.t.}}
\newcommand{\prox}{\mbox{prox}}

\newcommand{\argmin}{\mathop{\rm arg\;min}}

\newcommand{\R}{\mathds{R}}
\newcommand{\E}{\mathbb{E}}

\newcommand{\summ}{\sum\limits_{n=1}^N}
\newcommand{\sumq}{\sum\limits_{n=1}^Q}

\newcommand{\sumam}{\sum\limits_{n=0}^N}
\newcommand{\sumkt}{\sum\limits_{k=1}^T}
\newcommand{\sumk}{\sum\limits_{k=1}^K}
\newcommand{\sumi}{\sum\limits_{i=1}^I}

\renewcommand{\u}{\bm u}
\renewcommand{\v}{\bm v}
\newcommand{\x}{\bm x}

\newcommand{\w}{\bm w}
\renewcommand{\tt}{\mbox{\tiny T}}

\newtheorem{theorem}{\textbf{Theorem}}
\newtheorem{lemma}{\textbf{Lemma}}

\newtheorem{remark}{\textbf{Remark}}
\newtheorem{assumption}{\textbf{Assumption}}

\begin{document}
\title{Communication-Efficient Byzantine-Resilient Federated Learning Over Heterogeneous Datasets}

\author{
	\mbox{Yanjie Dong,~\IEEEmembership{Student Member, IEEE},}
	\mbox{Georgios B. Giannakis,~\IEEEmembership{Fellow, IEEE},}
	\mbox{Tianyi Chen,~\IEEEmembership{Member, IEEE},}
	\mbox{Julian Cheng,~\IEEEmembership{Senior Member, IEEE},}
	\mbox{Md. Jahangir Hossain,~\IEEEmembership{Senior Member, IEEE},}
	\mbox{and Victor C. M. Leung~\IEEEmembership{Fellow, IEEE},}

\thanks{
This work was supported in part by a Mitacs Globalink Research Award, in part by a UBC Four-Year Doctoral Fellowship, in part by the Natural Science and Engineering Research Council of Canada, and in part by the National Engineering Laboratory for Big Data System Computing Technology at Shenzhen University, China. \emph{(Corresponding author: Victor C. M. Leung.)}}
\thanks{Y.~Dong is with the Department of Electrical and Computer Engineering, The University of British Columbia, Vancouver, BC V6T 1Z4, Canada (email: ydong16@ece.ubc.ca).}
\thanks{G.~B.~Giannakis is  with the Department of Electrical and Computer Engineering and the Digital Technology Center, University of
Minnesota, Minneapolis, MN 55455, USA (email: georgios@umn.edu).}
\thanks{T.~Chen is with the Department of Electrical, Computer, and Systems Engineering, Rensselaer Polytechnic Institute, Troy, NY 12180, USA (email: chent18@rpi.edu).}
\thanks{J.~Cheng and M.~J.~Hossain are with the School of Engineering, The University of British Columbia, Kelowna, BC V1V 1V7, Canada (email: \{julian.cheng, jahangir.hossain\}@ubc.ca).}
\thanks{V.~C.~M.~Leung is with the College of Computer Science and Software Engineering, Shenzhen University, Shenzhen 518060, China, and the Department of Electrical and Computer Engineering, The University of British Columbia, Vancouver, BC \mbox{V6T 1Z4}, Canada (e-mail: vleung@ieee.org).}
}

\maketitle
\begin{abstract}
	This work investigates fault-resilient federated learning when the data samples are non-uniformly distributed across workers, and the number of faulty workers is unknown to the central server. In the presence of adversarially faulty workers who may strategically corrupt datasets, the local messages exchanged (e.g., local gradients and/or local model parameters) can be unreliable, and thus the vanilla stochastic gradient descent (SGD) algorithm is not guaranteed to converge. Recently developed algorithms improve upon vanilla SGD by providing robustness to faulty workers at the price of slowing down convergence. 
	To remedy this limitation, the present work introduces a fault-resilient proximal gradient (FRPG) algorithm that relies on Nesterov's acceleration technique. To reduce the communication overhead of FRPG, a local (L) FRPG algorithm is also developed to 
	allow for intermittent server-workers parameter exchanges. For strongly convex loss functions, FRPG and LFRPG have provably faster convergence rates than a benchmark robust stochastic aggregation  algorithm. 
	Moreover, LFRPG converges faster than FRPG while using the same communication rounds. 
	Numerical tests performed on various real datasets confirm the accelerated convergence of FRPG and LFRPG over the robust stochastic aggregation benchmark and competing alternatives. 
\end{abstract}

\section{Introduction}
Traditional machine learning algorithms are mostly designed for centralized processing of datasets at a single server or a cloud. However, the copies of datasets in a cloud render cloud-centric learning vulnerable to privacy leakage \cite{Li2019a, Li2019b, Konecny2015}. 
Therefore,  distributed on-device learning has emerged to alleviate these privacy concerns by using the ever-improving computational capability of network-edge devices, such as edge base stations, mobile terminals, and Internet of Things (IoT) devices.

As an implementation of distributed on-device learning, federated learning has attracted growing attention from both industry and academia \cite{Konecny2015, Dong2019, chen2019joint, 9134426}. A popular realization of federated learning employs a server collaborating with multiple workers, which are the network-edge devices in practical systems.  Specifically, a server updates and broadcasts global models\footnote{For example, models are employed to provide a mapping between data samples and labels in supervised learning.} using local messages (e.g., local gradients and local models) from workers.
Based on the received global models and local datasets, workers compute local messages in parallel. Since datasets are kept at workers in federated learning, the risk of privacy leakage is reduced. In the context of federated learning, recent research ranges from allocating resource at the physical layer \cite{chen2019joint} to design learning algorithms at the upper layers \cite{Dong2019}. The two major topics in the design of learning algorithms deal with fault-resilient and communication-efficient federated learning. The present work builds on this research front.  

\subsection{Related Works}
\textbf{Fault-resilient federated learning}. 
Faulty workers may strategically corrupt the local datasets in federated learning such that the local messages uploaded to the server are unreliable. When unreliable local messages are used by a server, the convergence of the vanilla gradient descent algorithm is not guaranteed. Indeed, it has been demonstrated  that the vanilla gradient descent algorithm and its stochastic version (Stochastic Gradient Descent, SGD) fail to converge when each faulty worker uploads an unreliable message to the server \cite{chen2017}. Therefore, it is crucial to deal with faulty workers. Fault-resilient learning approaches have been reported in recent research relying on the full gradient per update \cite{chen2017, pmlr-v80-yin18a, pmlr-v97-yin19a, su2019}. For strongly convex loss functions, the geometric median \mbox{(GeoMed)} algorithm \cite{chen2017} converges to a near-optimal solution when less than $50\%$ of the messages is from unreliable workers. 
For (strongly) convex and smooth \mbox{non-convex} loss functions, a \mbox{component-wise} median and \mbox{component-wise} trimmed mean algorithms \cite{pmlr-v80-yin18a} were developed to secure model updates at a server over faulty workers.  
For \mbox{non-convex} loss functions, the component-wise median and component-wise trimmed mean algorithms may converge to a saddle point that is far away from a real local minimizer. As a remedy, a Byzantine perturbed gradient algorithm \cite{pmlr-v97-yin19a} was proposed to obtain a solution that is an approximate local minimizer of the non-convex loss function. 

For large datasets, however, evaluating the full gradient per iteration is computationally prohibitive. 
Therefore, fault-resilient stochastic federated learning is proposed to improve computational efficiency such as Krum \cite{NIPS2017_6617}, Bulyan \cite{pmlr-v80-mhamdi18a}, Byzantine SGD \cite{Alistarh2018}, Zeno \cite{pmlr-v97-xie19b}, DRACO \cite{Chen2018}, and SLSGD \cite{xie_slsgd}. 
When less than $50\%$ of the workers are unreliable, the Krum algorithm can obtain the dissimilarity score of local gradients. Using the local gradient with the smallest dissimilarity score to update the global models, the Krum algorithm converges to a neighborhood of stationary points \cite{NIPS2017_6617}. 
When less than $25\%$ of the workers are unreliable, the Buylan algorithm \cite{pmlr-v80-mhamdi18a} can improve the accuracy of the suboptimal solution obtained by the Krum algorithm. 
Using historical gradients, the Byzantine SGD algorithm \cite{Alistarh2018} allows the server to remove the faulty local gradients before performing gradient aggregation. 
The Zeno algorithm \cite{pmlr-v97-xie19b} ranks the reliability of local gradients based on the weighted descent value and magnitude of local gradients. Averaging the top-ranked local gradients, the Zeno algorithm can tolerate up to $Q-1$ faulty workers, where $Q$ is the number of workers. Based on coding theory and sample redundancy (i.e., multiple copies of a data sample across different workers), the DRACO algorithm \cite{Chen2018} converges when there is at least one reliable worker. 

While the homogeneous datasets\footnote{Data samples from different workers are independent and identically distributed} are used to design learning algorithms in aforementioned works \cite{NIPS2017_6617, pmlr-v80-mhamdi18a, Alistarh2018, pmlr-v97-xie19b, Chen2018}, the datasets are non-independent or non-identically distributed across different workers (a.k.a. heterogeneous datasets) in several practical settings.
For example, different YouTube subscribers are provided with different categories of advertisements and video clips based on their search history. 
As a result, developing fault-resilient federated learning algorithms over heterogeneous datasets has emerged as an important research task. 
For heterogeneous datasets, a robust stochastic aggregation framework \cite{Li2019} was introduced to optimize fault-resilient stochastic federated learning. 
The resultant robust stochastic aggregation (RSA) algorithm can converge to a near-optimal solution with convergence rate $\OO{\nicefrac{\log K}{\sqrt K}}$, where $K$ is the number of iterations; see also \cite{Ghosh2019} where multi-task federated learning is effective for heterogeneous datasets over worker clusters having different models, but not effective for a single-model problem. 

\textbf{Communication-Efficient Federated Learning}. 
Frequent communications between the server and workers are inevitable in federated learning. 
Since bandwidth is a scarce resource for the server, the communication overhead becomes the bottleneck \cite{NIPS2014_5597, michaeljason2019, 8889996}. 
To reduce this overhead, a line of research focuses on skipping the unnecessary communication rounds, where the LAG algorithm avoids redundant information exchanges \cite{Chentobepublished}, and can be extended to employ just quantized gradients \cite{Sun2019}. 
Compared with the vanilla gradient descent algorithm, the LAG enjoys comparable convergence at reduced communication overhead; see also  
\cite{Stich2019, yu2018parallel, pmlr-v97-yu19d, khaled2019first} that leverage local SGD to allow intermittent server-worker exchanges. 
The  remaining unexplored issue is whether LAG and local SGD algorithms are robust to faulty workers. 

\subsection{Contributions}
Motivated by the need for \mbox{communication-efficient} robust learning over heterogeneous datasets, we propose two communication-efficient federated learning algorithms that are robust to faulty workers. Our contributions are as follows.
\begin{itemize}
	\item In the presence of faulty workers, heuristically using Nesterov's acceleration leads to divergence of the vanilla SGD algorithm. 
	As a remedy, we develop a fault-resilient proximal gradient (FRPG) algorithm by tailoring Nesterov's acceleration  \cite{Nesterov:2014:ILC:2670022, Hu2009} and stochastic approximation for fault-resilient stochastic federated learning. 
	\item To reduce communication overhead further, we also develop a local FRPG (LFRPG) algorithm where the server periodically communicates with workers, and we prove that LFRPG has lower communication overhead than FRPG.  
	\item We establish the convergence rates for the proposed FRPG and LFRPG algorithms, which are challenging to analyze when faulty workers are present. 
	Our theoretical results demonstrate that the proposed FRPG and LFRPG algorithms can converge faster than the existing federated learning schemes.
\end{itemize}

Numerical tests are performed over practical datasets, and corroborate our analytical findings.  

The remaining work is organized as follows. 
The investigated problem is described in Section II.  
The FRPG algorithm and its convergence analysis are the subjects of Section III, and the LFRPG algorithm and its convergence analysis are presented in Section IV. 
Numerical results are shown in Section V, and conclusions are drawn in Section VI. 

\emph{Notation.}
The $\ell_2$-norm of a vector is denoted by $\norm{\cdot}$.
The operator $\vec{w_1, w_2, \ldots, w_N}$ returns a column vector by stacking $w_1, w_2,\ldots, w_N$. 
The operator $\inp{\cdot, \cdot}$ denotes the inner product of two vectors. 
The operator $\E_{x}\sq{\cdot}$ denotes the expectation over the random variable $x$. 
The polynomial of  $x$ is denoted by $\OO{x}$.
The proximal operator for a function $f$ is defined as
\begin{equation*}
\prox_{\alpha f}\br{w} := \argmin\limits_{x}\cb{  \alpha f\br{x} + \frac{1}{2}\norm{x - w}^2 }.
\end{equation*}
The nomenclature of this work is listed in Table \ref{tab:my_label}. 

\begin{table}[ht]
    \centering
    \caption{Nomenclature}
    \label{tab:my_label}
    \begin{tabular}{cl}
    \toprule
       \textbf{Notations} & \textbf{Definitions} \\
    \midrule
        $Q$ & Number of workers \\
        $N$ & Number of reliable workers \\
        $B$ & Number of faulty workers \\
        $G$ & Maximum gradient power of penalty functions  \\
        $\lambda$ & Weight factor for penalty functions \\
        $f_0\br{w_0}$ & Regularization function at the server at $w_0$\\
        $f_n\br{w_n}$ & Local loss at the $n$th worker at $w_n$\\
        $f\br{w_n; x_n}$ & Loss value at the $n$th worker w.r.t.  random variable $x_n$\\
        $p_n\br{w_0 - w_n}$~ & Penalty function at the $n$th worker \\
        $\beta_k$ & Step size in the $k$th slot\\
        $\alpha_{0,k}$, $\alpha_{n,k}$ & Step sizes in the $k$th slot of server and the $n$th worker \\
        $u_{0,k}$, $v_{0,k}$ & Auxiliary sequences at the server in the $k$th slot \\
        $u_{n,k}$, $v_{n,k}$ & Auxiliary sequences at the $n$th worker in the $k$th slot \\
        $w_{0,k}$ & Model parameters at the server in the $k$th slot \\
        $w_{n,k}$ & Model parameters at the $n$th worker in the $k$th slot \\
        $g_{n,k}$ & Gradient of the $n$th penalty  $\lambda\nabla_{w_0}p_n\br{w_{0,k} - w_{n,k}}$\\
        $\Delta_{0,k}$ & Gradient noise at the server, i.e., $\sum\nolimits_{n=N+1}^Q g_{n,k}$ \\
        $\Delta_{n,k}$ & Gradient noise at the $n$th worker in the $k$th slot \\
        \bottomrule
    \end{tabular}
\end{table}

\section{Problem statement}
Consider a federated learning setup, comprising a parameter server, $Q$ workers, and overall loss given by \cite{Konecny2015}
\begin{equation}\label{eqa:pd:01}
\sumq f_n\br{w_n} + f_0\br{w_0}
\end{equation}
where $w_0 \in \R^d$ denotes model parameters at the server; $w_n \in \R^d$ are model parameters at the $n$th worker; and $f_0\br{w_0}$ is a regularization function.
The local loss at the $n$th worker is
\begin{equation}\label{eqa:pd:02}
f_n\br{w_n } = \E_{x_n}\sq{ f\br{w_n; x_{n}} }
\end{equation}
where $\E_{x_n}\sq{\cdot}$ denotes the $n$th worker's specific expectation over the random data vector $x_{n}$, and $f\br{w_n; x_n}$ is the corresponding loss with respect to $w_n$ and $x_n$.

The objective of fault-resilient federated learning is to minimize in a \emph{distributed} fashion the loss in \eqref{eqa:pd:01} subject to the consensus constraints, expressed as 
\begin{equation}\label{eqa:pd:03}
w_0 = w_n, ~~~~n = 1, \ldots, Q.
\end{equation}

When there are multiple faulty workers, several researchers have demonstrated that obtaining the minimizer of \eqref{eqa:pd:01} subject to \eqref{eqa:pd:03} is less meaningful \cite{NIPS2017_6617, chen2017, Li2019}. For this reason, our goal will be to minimize the loss function while avoiding consensus with faulty workers. The server cannot differentiate reliable from faulty workers, and does not even know the number of faulty workers. Our novel algorithms will seek resilience to faulty workers under these challenging conditions. But when analyzing the convergence rate in the presence of faulty workers, we will assume that among $Q$ workers, $N$  are reliable, and for notational convenience we will index the $B=Q-N$ faulty workers by $n = N+1, \ldots, Q$. 

Dropping the losses of  faulty workers in \eqref{eqa:pd:01}, the \emph{ideal} minimization task with 
$\w := \vec{w_1, w_2, \ldots, w_N}$, is
\begin{equation}\label{eqa:pd:04}
\begin{split}
\min\limits_{\w}&\; \summ f_n\br{w_n} + f_0\br{w_0} \\
\st&\; w_0 = w_n, ~~~~n = 1, \ldots, N\:.
\end{split}
\end{equation}
Without information about faulty workers, it is ideal (and thus not meaningful) for the server to seek the solution of \eqref{eqa:pd:04}. Instead, we will adapt the robust stochastic aggregation approach of \cite{Li2019} by adding a penalty term $p_n\br{w_0 - w_n}$ with weight $\lambda>0$ per local loss $f_n\br{w_n}$. We will then target to approach the solution of the penalized version of \eqref{eqa:pd:04}, namely
\begin{equation}\label{eqa:pd:06}
\min\limits_{\w}  F\br{\w} := \summ \br{f_n\br{w_n} + \lambda p_n\br{w_0 - w_n}} + f_0\br{w_0}\;.
\end{equation}

\begin{remark}
Different from \eqref{eqa:pd:04}, the penalty terms in \eqref{eqa:pd:06} allow the server model and those of faulty workers to differ. 
Besides, this flexibility is also preferred to handle data heterogeneity across workers \cite{KoppelJune2017}. 
Therefore, our proposed algorithms and corresponding convergence analysis are based on \eqref{eqa:pd:06} in this work. 
We will select convex and differentiable $\{p_n(\cdot)\}$, e.g., of the Huber type. 
Moreover, the gradients of 	$\{p_n(\cdot)\}$ for reliable and faulty workers must be similar, so that the undesirable influence of faulty workers is mitigated.
\end{remark}

Our communication-efficient solvers of a non-ideal version of \eqref{eqa:pd:06} will be developed in Sections III and IV, based on the following assumptions about $f_0$, $f_n$, and $p_n$, for $n = 1, \ldots, N$.
\begin{assumption}[Lipschitz Continuity {\cite[eq. (1.2.11)]{Nesterov:2014:ILC:2670022}}]\label{as:01}
Regularizer $f_0$ has an $L_0$-Lipschitz continuous gradient, and $f_n$ has an $L_n$-Lipschitz continuous gradient for $n = 1, \ldots, N$.
\end{assumption}

\begin{assumption}[Strong Convexity {\cite[eq. (2.1.20)]{Nesterov:2014:ILC:2670022}}]\label{as:02}
	Regularizer $f_0$ is strongly convex with modulus $\delta_0$, and loss $f_n$ is strongly convex with modulus $\delta_n$ for $n = 1, \ldots, N$. 
\end{assumption}

\begin{assumption}[Penalty]\label{as:03}
	Penalty function $p_n\br{w_0 - w_n}$ is convex and differentiable, with  $\|{\nabla_{w_{0}} p_n\br{w_0 - w_n}}\|^2 \le G$, and $\|{\nabla_{w_{n}} p_n\br{w_0 - w_n}}\|^2 \le G$ for $n = 1, \ldots, Q$. 
\end{assumption}

Note that Assumption \ref{as:01} is easily satisfied. Several functions have Lipschitz-continuous gradients, such as the square of $\ell_2$-norm, logistic regression function, and multinomial logistic regression function. Besides, some artificial neural networks also have Lipschitz-continuous gradients \cite{latorre2020lipschitz}. Assumption \ref{as:02} can also be easily satisfied when the square of $\ell_2$-norm is added to convex functions. Assumptions \ref{as:01} and \ref{as:02} are standard when the learning criterion entails smooth and strongly convex local loss functions. The negative effects of faulty workers can be bounded through Assumption \ref{as:03}, which is satisfied by, e.g., a Huber-type penalty.

\begin{figure}[t]
\centering
  \includegraphics[width=3.5 in]{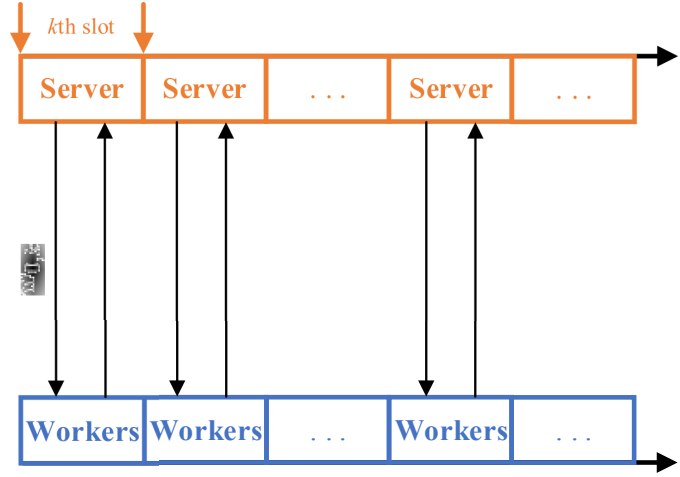}
  \caption{Per FRPG iteration, the server broadcasts $w_{0, k}$, and the workers upload $g_{n,k}$, $n = 1, \ldots, Q$. }\label{fg:brpg}
\end{figure}
\section{Fault-Resilient Proximal Gradient}
In this section, we develop a novel fault-resilient proximal gradient (FRPG) algorithm for the server to solve the \emph{non-ideal} version of \eqref{eqa:pd:06}, with $Q$ replacing $N$ since faulty workers can be present. Subsequently, we will analyze the convergence of our iterative FRPG solver.
\subsection{Algorithm} 
Along the lines of \cite{Hu2009}, the parameter server in our federated learning approach  maintains three sequences per slot $k$, namely $u_{0, k}$, $w_{0,k}$ and $v_{0, k}$. 
The resultant FRPG algorithm updates these three sequences using the  recursions 
\begin{subequations}\label{eqa:brpg:01}
\begin{align}
\hspace{-0.3 cm}
u_{0,k} &= \br{1-\beta_k}w_{0, k-1} + \beta_k v_{0, k-1} \label{eqa:brpg:01a}\\
\hspace{-0.3 cm}w_{0,k} &= u_{0, k} - \frac{1}{\alpha_{0, k}}\nabla f_0\br{u_{0, k}} \label{eqa:brpg:01b}\\
\hspace{-0.3 cm}v_{0,k} &= v_{0, k-1} - \frac{\delta_0\br{v_{0, k-1} - u_{0, k}} + \nabla f_0\br{u_{0, k}} + \sumq g_{n,k}}{\delta_0 + \alpha_{0,k}\beta_k} \label{eqa:brpg:01c}
\end{align}
\end{subequations}
where $w_{0, k-1}$ are the server parameters on slot \mbox{$\br{k-1}$}; and likewise for the 
auxiliary iterates $v_{0, k-1}$; scalars $\alpha_{0, k}$ and $\beta_k$ are step sizes; and the sum over $Q$ in \eqref{eqa:brpg:01c} accounts for the non-ideal inclusion of faulty workers, where $g_{n, k}$ is given by  
\begin{equation}\label{eqa:brpg:01d}
    g_{n, k} :=  \lambda\nabla_{w_0}p_n\br{w_{0,k} - w_{n,k}}.
\end{equation}
Each reliable worker also maintains sequences $u_{n, k}$, $w_{n,k}$ and $v_{n, k}$
per slot $k$, that are locally updated as  
\begin{subequations}\label{eqa:brpg:02}
\begin{align}
\hspace{-0.25 cm}u_{n,k} &= \br{1-\beta_k}w_{n, k-1} + \beta_k v_{n, k-1} \label{eqa:brpg:02a}\\
\hspace{-0.25 cm}w_{n,k} &= w_{0, k} - \prox_{\!\frac{\lambda p_n}{\alpha_{n, k}}}\!\!\cb{\!w_{0, k} \!-\! u_{n, k}
+ \frac{\nabla\! f\br{u_{n,k}; x_{n,k}}}{\alpha_{n, k}} \!} \label{eqa:brpg:02b}\\
\hspace{-0.25 cm}v_{n,k} &= v_{n,k-1} \nonumber\\
&\hspace{0.5 cm} - \frac{\delta_n\br{v_{n, k-1} - u_{n,k}} + \nabla f\br{u_{n,k}; x_{n,k}} - g_{n,k}}{\delta_n + \alpha_{n,k}\beta_k} \label{eqa:brpg:02c}
\end{align}
\end{subequations}
where subscript $k-1$ indices the previous slot; while $\alpha_{n,k}$ and $\beta_k$ denote stepsizes as before; and $x_{n,k}$ is a realization of $x_{n}$ at slot $k$. Without adhering to \eqref{eqa:brpg:02a}-\eqref{eqa:brpg:02c}, faulty workers generate parameters $\{w_{n,k}\}_{n=N+1}^Q$ using an unknown mechanism. 

Based on \eqref{eqa:brpg:01} and \eqref{eqa:brpg:02}, our novel FRPG solver of \eqref{eqa:pd:06} is listed under Algorithm \ref{alg:01} with lines \ref{line:04}--\ref{line:10} showing that the workers generate their local models in parallel. The motivations of several key steps in the FRPG solver are as follows.
\begin{itemize}
	\item After receiving $g_{n,k}$, the server performs summation over all penalty gradients $\{g_{n,k}\}_{n=1}^Q$ as shown in \eqref{eqa:brpg:01c}. Since the server has no information on faulty workers, the negative effects of faulty workers are mitigated by using bounded penalty gradients $\{\nabla_{w_n} p_n(w_0 - w_n)\}_{n=1}^Q$. 
	More specifically, we observe from the term $\sum_{n=1}^Q g_{n,k}$ in \eqref{eqa:brpg:01c} that the impacts of a reliable worker and a faulty worker on $v_{0,k}$ are similar. 
	Recalling the bounded gradient property of penalty, we envision that the number of faulty workers (instead of the magnitudes of faulty model $\{w_{n,k}\}_{n = N+1}^Q$) will have influence on the update in \eqref{eqa:brpg:01c}. In this case, the FRPG algorithm is robust to any type of faulty workers.
	\item After receiving $w_{0,k}$, each reliable worker $n$ performs the local calculation \eqref{eqa:brpg:02}. In the presence of faulty workers, it is reasonable to allow a slight difference between the reliable models $\{w_{n,k}\}_{n=1}^{N}$ and the server model $w_{0,k}$ at slot $k$. Therefore, the proximal step in \eqref{eqa:brpg:02b} is used to obtain the reliable model $w_{n,k}$ while retaining a slight difference from the server model $w_{0,k}$, $n = 1,\ldots, N$. Besides, the local model $w_{n,k}$ is updated based on $\nabla_{w_n} p_n(w_{0,k} - w_{n,k})$ when proximal step  \eqref{eqa:brpg:02b} is used; otherwise, the local model $w_{n,k}$ is updated based on outdated information $\nabla_{w_n} p_n(w_{0,k} - w_{n,k-1})$ that slows down the convergence.
\end{itemize}

\begin{remark}
    Note that while our algorithm is inspired by \cite{Hu2009}, the updates in \eqref{eqa:brpg:01} and \eqref{eqa:brpg:02} are distinct in three aspects. The update step in \eqref{eqa:brpg:01b} does not require a proximal operation since $w_{0,k}$ and $w_{n,k}$ must be iteratively updated. Since FRPG is a distributed algorithm, the update steps in \eqref{eqa:brpg:01c} and \eqref{eqa:brpg:02c} require server-worker exchanges of $\{g_{n,k}\}_{n=1}^Q$ that also include exchanges from faulty workers. These three differences render the ensuing convergence analysis of FRPG challenging.  
\end{remark} 

\begin{algorithm}[t]\small
  \centering
  \caption{FRPG Algorithm}\label{alg:01}
  \begin{algorithmic}[1]
  \State \textbf{Initialize}: $w_{n, 0}$ and $v_{n,0}$ for $n = 0,\ldots,N$, and step sizes as \eqref{eqa:brpg:12} and \eqref{eqa:brpg:13}
  \For{$k = 1, \ldots, K$}
  \State The server updates $u_{0,k}$ and $w_{0,k}$ via \eqref{eqa:brpg:01a} and \eqref{eqa:brpg:01b}
  \State The server broadcasts the model parameters $w_{0,k}$
  \ParFor{$n = 1, \ldots, Q$} \label{line:04}\Comment{Parallel Computation}
  \If{$n = 1, \ldots, N$}
  \State The $n$th reliable worker updates $w_{n, k}$ via \eqref{eqa:brpg:02}
  \EndIf
  \If{$n = N+1, \ldots, Q$}
  \State The $n$th faulty worker generates faulty parameters
  \EndIf
  \EndParFor\label{line:10}
  \State All workers upload $g_{n, k}$ to the server
  \State The server updates $v_{0,k}$ via \eqref{eqa:brpg:01c}
  \EndFor
  \end{algorithmic}
\end{algorithm}

\subsection{Convergence analysis}
Our analysis here is for a single realization of $x_n$ per slot, but can be directly extended to mini-batch realizations of $x_n$. Let us define the gradient error at worker $n$ per slot $k$ as 
\begin{equation}\label{eqa:brpg:03}
\Delta_{n,k} := \nabla f\br{u_{n,k}; x_{n,k}} - \nabla f_n\br{u_{n,k}}
\end{equation}
and adopt the following assumption on its moments that are satisfied, e.g., when stochastic gradients are employed~\cite{Nemirovski2009}.

\begin{assumption}[Bounded Stochastic Noise]\label{as:04}
	The gradient error (a.k.a. noise) is zero mean, that is $\E_{x_n}[\Delta_{n,k}] = 0$, with bounded variance $\E_{x_n}[{\|{\Delta_{n,k} }\|^2}] \le \sigma_n^2$, for $n = 1, \ldots, N$.
\end{assumption}

\begin{lemma}\label{le:01}
If Assumptions \ref{as:01}--\ref{as:03} hold, \eqref{eqa:brpg:01b} implies that 
\begin{align}
& f_0\br{w_{0,k} } - f_0\br{u_0} \label{eqa:brpg:04}\\
\le&  \inp{ \summ g_{n,k} + \Delta_{0,k}, u_0 \!-\! w_{0,k}} 
\! - \! \br{\alpha_{0,k} - \frac{L_0}{2}}\norm{u_{0, k}-w_{0,k}}^2 \nonumber \\
& + \norm{\sumq g_{n,k}}\norm{u_{0, k}-w_{0,k}} 
- \frac{\delta_0}{2}\norm{u_0 - u_{0,k}}^2 \nonumber\\
& - \inp{\alpha_{0,k}\br{u_{0,k}-w_{0,k}}+\sumq g_{n,k}, u_0 - u_{0,k}}\nonumber
\end{align}
where $\Delta_{0,k} := \sum_{n=N+1}^Q g_{n,k}$, and $u_0 \in\R^d$ is an arbitrary vector. 
\end{lemma}
\begin{IEEEproof}
See Appendix \ref{apdx:01}.
\end{IEEEproof}

\begin{lemma}\label{le:02}
If Assumptions \ref{as:01}--\ref{as:03} hold, \eqref{eqa:brpg:02b} implies that  
\begin{align}\label{eqa:brpg:05}
& f_n\br{w_{n,k}} - f_n\br{u_n} \\
\le & \inp{\Delta_{n,k} - g_{n,k}, u_n - w_{n,k}}
 - \br{\alpha_{n,k} - \frac{L_n}{2}}\norm{u_{n,k}-w_{n,k}}^2 \nonumber\\
& - \alpha_{n,k}\inp{{u_{n,k}-w_{n,k}}, u_n - u_{n,k}}
- \frac{\delta_n}{2}\norm{u_n - u_{n,k}}^2  \nonumber
\end{align}
where $u_n \in \R^d$ is an arbitrary vector, $n = 1, \ldots, N$. 
\end{lemma}
\begin{IEEEproof}
See Appendix \ref{apdx:02}.
\end{IEEEproof}

Since $p_n\br{w_0 - w_n}$ is convex and differentiable (cf. Assumption 3), eq.~\eqref{eqa:brpg:01d} implies that $\lambda\nabla_{w_n}p\br{w_{0,k} - w_{n,k}} = -g_{n,k}$, and thus 
\begin{equation}\label{eqa:brpg:06}
\begin{split}
&\lambda p_n\br{w_{0,k} - w_{n,k}} - \lambda p_n\br{u_0 - u_n}  \\
\le& \inp{ g_{n,k}, u_n - w_{n,k}} -  \inp{ g_{n,k}, u_0 - w_{0,k}}. 
\end{split}
\end{equation}

Summing up \eqref{eqa:brpg:04}--\eqref{eqa:brpg:06} and using the definition of $F\br{\w}$ in \eqref{eqa:pd:06}, we obtain
\begin{align}
& F\br{\w_k} -  F\br{\u} \label{eqa:brpg:07}\\
\le & \sumam \inp{\Delta_{n,k}, u_n - w_{n,k}}  +
\norm{\sumq g_{n,k}}\norm{u_{0,k} - w_{0,k}} \nonumber\\
& - \sumam \br{\alpha_{n,k} - \frac{L_n}{2}}\norm{u_{n,k} - w_{n,k}}^2
- \sumam \frac{\delta_n}{2}\norm{u_n - u_{n,k}}^2 \nonumber\\
& - \summ\alpha_{n,k}\inp{u_{n,k} - w_{n,k}, u_n - u_{n,k}} \nonumber\\
& - \inp{\alpha_{0,k}\br{u_{0,k} - w_{0,k}} +  \sumq g_{n,k}, u_0 - u_{0, k}} \nonumber
\end{align}
where 
$\w_k := {\rm vec} ( w_{0, k}, \ldots, w_{N,k} )$, and 
$\u := {\rm vec} ( u_0,\ldots, u_n ) $.

Based on the definition of $\Delta_{0,k}$ and \mbox{Assumption \ref{as:03}}, it follows that $\|{\Delta_{0,k}}\| \le B\|{g_{1, k}}\|$ and $\|{\sum\nolimits_{n=1}^Q g_{n,k}}\| \le Q\|{ g_{1,k}}\|$. Using also that $\|{ g_{1, k}}\|^2 \le \lambda^2 G$,  $\|{\sum\nolimits_{n=1}^Q g_{n,k}}\| \le Q\|{ g_{1,k}}\|$ and $\|{\Delta_{0,k}}\| \le B\|{ g_{1, k}}\|$, we deduce that 
\begin{equation}\label{eqa:brpg:08}
\br{\norm{\sumq g_{n,k}} + \norm{\Delta_{0,k}}}^2 \le \lambda^2\br{Q+B}^2G := \sigma_0^2.
\end{equation}

\begin{lemma}\label{le:03}
Under Assumptions \ref{as:01}--\ref{as:04}, the FRPG iterates at the server relative to the optimum $\u^*$ satisfy  
\begin{equation}\label{eqa:brpg:09}
\begin{split}
& F\br{\w_k} -  F\br{\u^*}  \\
\le& \br{1-\beta_k}\br{F\br{\w_{k-1}} -  F\br{\u^*}}
+ \sumam \br{ \eta_{5, n, k} + \eta_{6, n, k} } \\
& + \frac{2\lambda^2Q^2 G}{\alpha_{0,k}}
+ \frac{\lambda^2 B^2 G}{2\epsilon}\beta_k
\end{split}
\end{equation}
where $\epsilon >0$, while the scalars  $\eta_{5, n, k}$ and $\eta_{6, n, k}$ are specified in   \eqref{apdx3:22} and \eqref{apdx3:23}, respectively.  
\end{lemma}
\begin{IEEEproof}
See Appendix \ref{apdx:03}.
\end{IEEEproof}

Using Lemma \ref{le:03}, our FRPG convergence is asserted next.  

\begin{theorem}[Convergence of FRPG]\label{th:01}
If under Assumptions \ref{as:01}--\ref{as:04}, the stepsizes are updated as
\begin{equation}\label{eqa:brpg:12}
\alpha_{n,k} =
\left\{\begin{array}{l}
\frac{\delta_0}{14}\br{k+2}^2 + \frac{3}{2}L_0, n = 0\\
\frac{3\delta_n}{14}\br{k+2}^2 + L_n, n = 1, \ldots, N
\end{array}\right.
\end{equation}
and
\begin{equation}\label{eqa:brpg:13}
\beta_k = \frac{2}{k+2} 
\end{equation}
FRPG converges as
\begin{equation}\label{eqa:brpg:14}
\begin{split}
 F\br{\w_k} - F\br{\u^*}  
\le & \frac{4}{\br{K+2}^2}\br{F\br{\w_0} - F\br{\u^*}+ \sumam \eta_{9, n} } \\
& + \frac{4 K }{\br{K+2}^2} \sumam\eta_{10, n} + \OO{\frac{\lambda^2B^2 G}{\delta_0}}
\end{split}
\end{equation}
where $K$ is the number of communication rounds, while scalars $\eta_{9, n}$ and $\eta_{10, n}$  are defined in \eqref{apdx4:08} and \eqref{apdx4:09}, respectively.
\end{theorem}
\begin{IEEEproof}
	See Appendix \ref{apdx:04}.
\end{IEEEproof}

As confirmed by the last term in \eqref{eqa:brpg:14}, FRPG converges to a neighborhood of the optimum with radius on the same order as that of RSA \cite{Li2019}, with rate $\OO{\nicefrac{1}{K^2} + \nicefrac{1}{K}}$, which is faster than $\OO{\nicefrac{\log K}{\sqrt{K}}}$ of RSA. This implies that FRPG is more communication-efficient than RSA. While achieving a faster convergence rate, FRPG still requires the workers to communicate with the parameter server on each slot. Our LFRPG algorithm developed in the next section reduces this overhead by skipping several communication rounds. 

However, two questions remain: i) what is the convergence rate of LFRPG? and, ii) how does the convergence of LFRPG depend on the communication period between the workers and parameter server? We answer these two questions next. 

\section{Local Fault-Resilient Proximal Gradient}
\begin{figure}[t]
	\centering
	\includegraphics[width = 3.5 in]{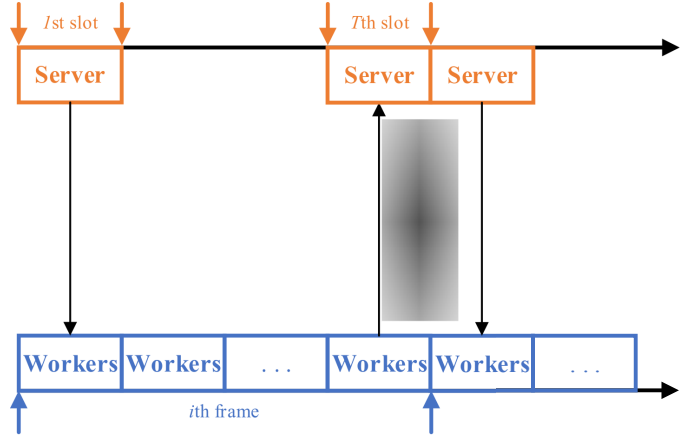}
	\caption{LFRPG iteration, where the server broadcasts $w_{0}^i$ at the beginning of the $i$th frame, and the workers upload $T^{-1}\sum\nolimits_{k=1}^T {g}_{n,k}^i$ at the end of the $i$th frame, $n = 1, \ldots, Q$.}\label{fg:lbrpg}  
	\vspace{-0.3cm}
\end{figure}

%\subsection{Algorithm}
To reduce the communication overhead, the model parameters at the server $w_0^i$ and the step sizes $\alpha_{n}^i$ and $\beta^i$ are updated at the start of the $i$th frame, $n = 0, 1, \ldots, N$ (as shown in Fig. \ref{fg:lbrpg}), with each frame consisting of $T$ slots.
The model parameters at the workers are updated in every slot. With $u_0^i$, $w_0^i$, and $v_0^i$ denoting the server sequences per frame $i$, the model parameters at the server are updated as 
\begin{subequations}\label{eqa:lbrpg:01}
	\begin{align}
	u_0^i &= \br{1-\beta^i} w_0^{i-1} + \beta^i v_0^{i-1} \label{eqa:lbrpg:01a}\\
	w_0^i &= u_0^{i} - \frac{1}{\alpha_{0}^{i}}\nabla f_0\br{u_0^i} \label{eqa:lbrpg:01b}\\
	v_0^i &= v_0^{i-1} - \frac{\delta_0\br{v_0^{i-1} - u_0^{i}} + \nabla f_0\br{u_0^i} + \frac{1}{T}\sumkt \sumq g_{n,k}^{i}}
	{\delta_0 + \alpha_{0}^{i}\beta^{i}}  \label{eqa:lbrpg:01c}
	\end{align}
\end{subequations}
where superscripts $i$ and $i-1$ index the corresponding frame in the sequences and stepsizes $\beta^i,\alpha_{0}^i$; while $g_{n,k}^i$ is defined as 
\begin{equation}\label{eqa:lbrpg:01d}
g_{n,k}^i := \lambda\nabla_{w_0}p_n({w_0^i - w_{n,k}^i}). 
\end{equation}
Accordingly, sequences at reliable worker $n$, slot $k$, and frame $i$ are updated using stepsizes 
$\alpha_{n}^{i},\beta^i$,  as 
\begin{subequations}\label{eqa:lbrpg:02}
\begin{align}
	u_{n,k}^{i} &= \br{1-\beta^i} w_{n,k}^{i-1} + \beta^i v_{n,k-1}^{i} \label{eqa:lbrpg:02a}\\
	w_{n,k}^{i} &= w_0^{i}
	- \prox_{\frac{\lambda p_n}{\alpha_{n}^{i}}}\!\cb{\! w_{0}^{i}
		- u_{n,k}^i + \frac{\nabla f(u^i_{n,k}; x^i_{n,k})}{\alpha_{n}^{i}} \!} \label{eqa:lbrpg:02b}\\
	v_{n,k}^i &= v_{n,k-1}^i \nonumber\\
	&\hspace{0.2 cm} - \frac{\delta_n(v_{n,k-1}^i - u_{n,k}^i) + \nabla f(u^i_{n,k}; x^i_{n,k}) -  g_{n,k}^i}{\delta_m + \alpha_{n}^i\beta^i} \label{eqa:lbrpg:02c}
\end{align}
\end{subequations}
while the resultant gradient noise is given by 
\begin{equation}\label{eqa:lbrpg:02d}
\Delta_{n,k}^i := \nabla f(u^i_{n,k}; x^i_{n,k}) - \nabla f_n(u_{n,k}^i).   
\end{equation}
Based on \eqref{eqa:lbrpg:01} and \eqref{eqa:lbrpg:02}, our novel scheme that we abbreviate as LFRPG, is listed in Algorithm \ref{alg:02}, where lines \ref{line2:04}--\ref{line2:10} show that the workers update local model parameters in parallel. To proceed with convergence analysis of LFRPG, we need an assumption on the per-frame gradient noise too. 

\begin{algorithm}[t]\footnotesize
	\centering
	\caption{LFRPG Algorithm}\label{alg:02}
	\begin{algorithmic}[1]
		\State \textbf{Initialize}: $w_{0}^0$, $v_{0}^0$, $w_{n,1}^0$ and $v_{n,0}^1$ for $n = 1,\ldots,N$, and step sizes as \eqref{eqa:lbrpg:09} and \eqref{eqa:lbrpg:10}. 
		\For{$i = 1, \ldots, I$}
		\State The server updates $u_{0}^i$ and $w_{0}^i$ via \eqref{eqa:lbrpg:01a} and \eqref{eqa:lbrpg:01b}
		\State The server broadcasts the model parameters $w_0^i$
		\For{$k = 1, \ldots, T$} \Comment{Local Iterations}
		\ParFor{$n = 1, \ldots, Q$}\label{line2:04} \Comment{Parallel Computation}
		\If{$n = 1, \ldots, N$}
		\State The $n$th reliable worker updates $w_{n,k}^i$ via \eqref{eqa:lbrpg:02}
		\EndIf
		\If{$n = N+1, \ldots, Q$}
		\State The $n$th faulty worker generates faulty parameters
		\EndIf
		\EndParFor\label{line2:10}
		\EndFor
		\State All workers upload $\frac{1}{T}\sum\nolimits_{k=1}^T {g}_{n,k}^i$ to the server
		\State The server updates $v_0^i$ via \eqref{eqa:lbrpg:01c}
		\EndFor
	\end{algorithmic}
\end{algorithm}

\begin{assumption}[Bounded Stochastic Noise]\label{as:05}
	The gradient noise is zero mean; that is, $\E_{x_n}[\Delta^i_{n,k}] = 0$, with bounded mean-square error: $\E_{x_n}[{\|{\Delta^i_{n,k} }\|^2}] \le \sigma_n^2$, for $n = 1, \ldots, N$.
\end{assumption}

\begin{lemma}\label{le:04}
	Under Assumptions \ref{as:01}--\ref{as:03} and \ref{as:05}, the descent loss at the server implied by the LFRPG iterates in \eqref{eqa:lbrpg:01b}, satisfies 
	\begin{align}
	& {f_0(w_0^{i}) - f_0(u_0)} \label{eqa:lbrpg:03}\\
	\le & \inp{ \summ g_{n,k}^i + \Delta_{0,k}^i, u_0 -  w_0^{i}}
	- \br{\alpha_0^i - \frac{L_0}{2}}\norm{u_0^i -  w_0^i}^2 \nonumber \\
	& + \norm{\sumq g_{n,k}^i}\norm{u_0^i - w_0^i} 
	- \frac{\delta_0}{2}\norm{u_0 - u_0^{i}}^2
	\nonumber\\
	& - \inp{\alpha_0^i\br{ u_0^i- w_0^i}+\sumq g_{n,k}^i,  u_0 -  u_0^i} \nonumber
	\end{align}
	where $\Delta_{0,k}^i := \sum\nolimits_{n=N+1}^Q g_{n,k}^i$.
\end{lemma}
\begin{IEEEproof}
	The proof follows directly from Lemma \ref{le:01}.
\end{IEEEproof}

\begin{lemma}\label{le:05}
	Under Assumptions \ref{as:01}--\ref{as:03} and \ref{as:05}, the descent loss per worker 
	implied by LFRPG iterates in \eqref{eqa:lbrpg:02b}, obeys
	\begin{align}
	& {f_n(w_{n,k}^i) - f_n(u_n)} \label{eqa:lbrpg:04}\\
	\le & \inp{ \Delta_{n,k}^i - g_{n,k}^i, u_n - w_{n,k}^i}
	- \br{\alpha_n^i - \frac{L_n}{2}}\norm{u_{n,k}^i - w_{n,k}^i}^2 \nonumber\\
	& - \alpha_n^i\inp{u_{n,k}^i - w_{n,k}^i, u_n - u_{n,k}^i}
	- \frac{\delta_n}{2}\norm{u_n - u_{n,k}^i}^2. \nonumber
	\end{align}
\end{lemma}
\begin{IEEEproof}
	The proof follows readily from Lemma \ref{le:02}.
\end{IEEEproof}

Since $u_0^i$ and $w_0^{i}$ are updated at the start of frame $i$, we set $u_0^i = u_{0,k}^i$ and $w_0^{i} = w_{0,k}^{i}$ with $k = 1, \ldots, T$. Summing \eqref{eqa:lbrpg:03} and \eqref{eqa:lbrpg:04}, it follows after straightforward algebraic manipulations that the overall loss at $w_k^i := \vec{w_{0, k}^i, w_{1, k}^i, \ldots, w_{N, k}^i}$, obeys
\begin{align}
& F(\w_k^i) - F\br{\u} \label{eqa:lbrpg:05}\\
\le & \sumam\inp{\Delta_{n,k}^i,  u_n -  w_{n,k}^{i}}
+ \norm{\sumq g_{n,k}^i}\norm{ u_0^i -  w_0^i} \nonumber\\
& - \sumam\br{\alpha_n^i - \frac{L_n}{2}}\norm{ u_0^i -  w_0^i}^2
- \sumam\frac{\delta_n}{2}\norm{ u_n -  u_{n,k}^i}^2 \nonumber\\
& - \summ\alpha_{n,k}^i\inp{ u_{n,k}^i- w_{n,k}^i,  u_n - u_{n,k}^i} \nonumber\\
& - \inp{\alpha_n^i\br{ u_0^i -  w_0^i} + \sumq g_{n,k}^i, u_0 - u_{0, k}^{i}}. \nonumber
\end{align}

\begin{lemma}\label{le:06}
	Under Assumptions \ref{as:01}--\ref{as:03} and \ref{as:05}, LFRPG iterates incur loss relative to the optimum $\u^*$, that is bounded by
	\begin{align}\label{eqa:lbrpg:06}
	& \frac{1}{T}\sumkt F\br{\w_k^i} - F\br{\u^*} \le  \br{1-\beta^{i}}\br{ \frac{1}{T}\sumkt F\br{\w_k^{i-1}} - F\br{\u^*} }\nonumber \\
	&+ \frac{\lambda^2B^2 G }{2\epsilon} \beta^i   + \sumam \br{\beta^i}^2\br{ \eta_{14, n}^i + \eta_{15,n}^i }  
	\end{align}
	where $\eta_{14, n}^i$ and $\eta_{15, n}^i$  are defined in \eqref{apdx5:23}, and \eqref{apdx5:24}, respectively.
\end{lemma}
\begin{IEEEproof}
	See Appendix \ref{apdx:05} of the supplementary material.
\end{IEEEproof}

Lemma \ref{le:06} leads to the convergence result for LFRPG. 

\begin{theorem}[Convergence of LFRPG]\label{th:02}
	If Assumptions \ref{as:01}--\ref{as:03} and \ref{as:05} hold, and stepsizes are respectively updated as
	\begin{equation}\label{eqa:lbrpg:09}
	\alpha_{n}^{i} =
	\left\{\begin{array}{l}
	\frac{\delta_0}{14}\br{i+2}^2 + \frac{3}{2}L_0, ~~~~n = 0\\
	\frac{3\delta_n}{14}\br{i+2}^2 + L_n, ~~~~n = 1, \ldots, N
	\end{array} \right.
	\end{equation}
	and
	\begin{equation}\label{eqa:lbrpg:10}
	\beta^i = \frac{2}{i+2}
	\end{equation}
	then average LFRPG iterates $\bar\w^I := T^{-1} \sum\nolimits_{k=1}^T\w_k^I$ converge
	\begin{equation}\label{eqa:lbrpg:11}
	\begin{split}
	\hspace{-0.2 cm}
	F\br{\bar\w^I} - F\br{\u^*} 
	\le  \frac{2 \eta_{16} }{T\br{I+2}^2}
	+ \frac{ \eta_{17}+ I\eta_{18} }{\br{I+2}^2}
	+ \OO{\frac{\lambda^2B^2 G}{\delta_0}}
	\end{split}
	\end{equation}
	where $\eta_{16}$, $\eta_{17}$ and $\eta_{18}$ are defined in \eqref{apdx6:05a}, \eqref{apdx6:05b} and \eqref{apdx6:05c}, respectively. 
\end{theorem}
\begin{IEEEproof}
	See Appendix \ref{apdx:06}. 
\end{IEEEproof}

The first fraction in \eqref{eqa:lbrpg:11} reveals that LFRPG outperforms FRPG in communication efficiency; while the last fraction asserts that LFRPG converges to the neighborhood of FRPG. 

\begin{figure*}[t]
	\centering
	\begin{minipage}{.3\textwidth}
		\includegraphics[width= 5.5 cm]{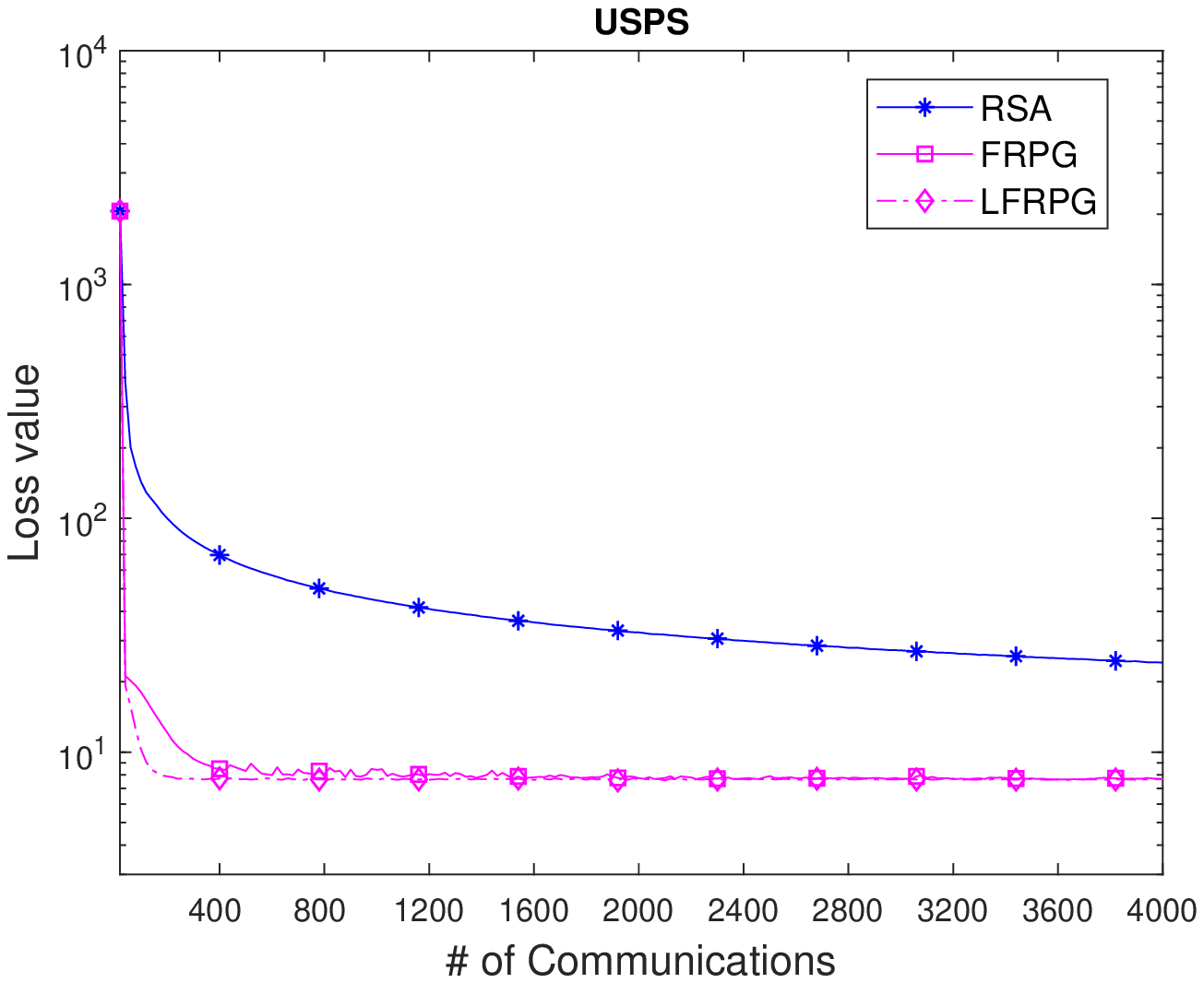}
	\end{minipage}%
\hspace{0.2 cm}
	\begin{minipage}{.3\textwidth}
		\includegraphics[width= 5.5 cm]{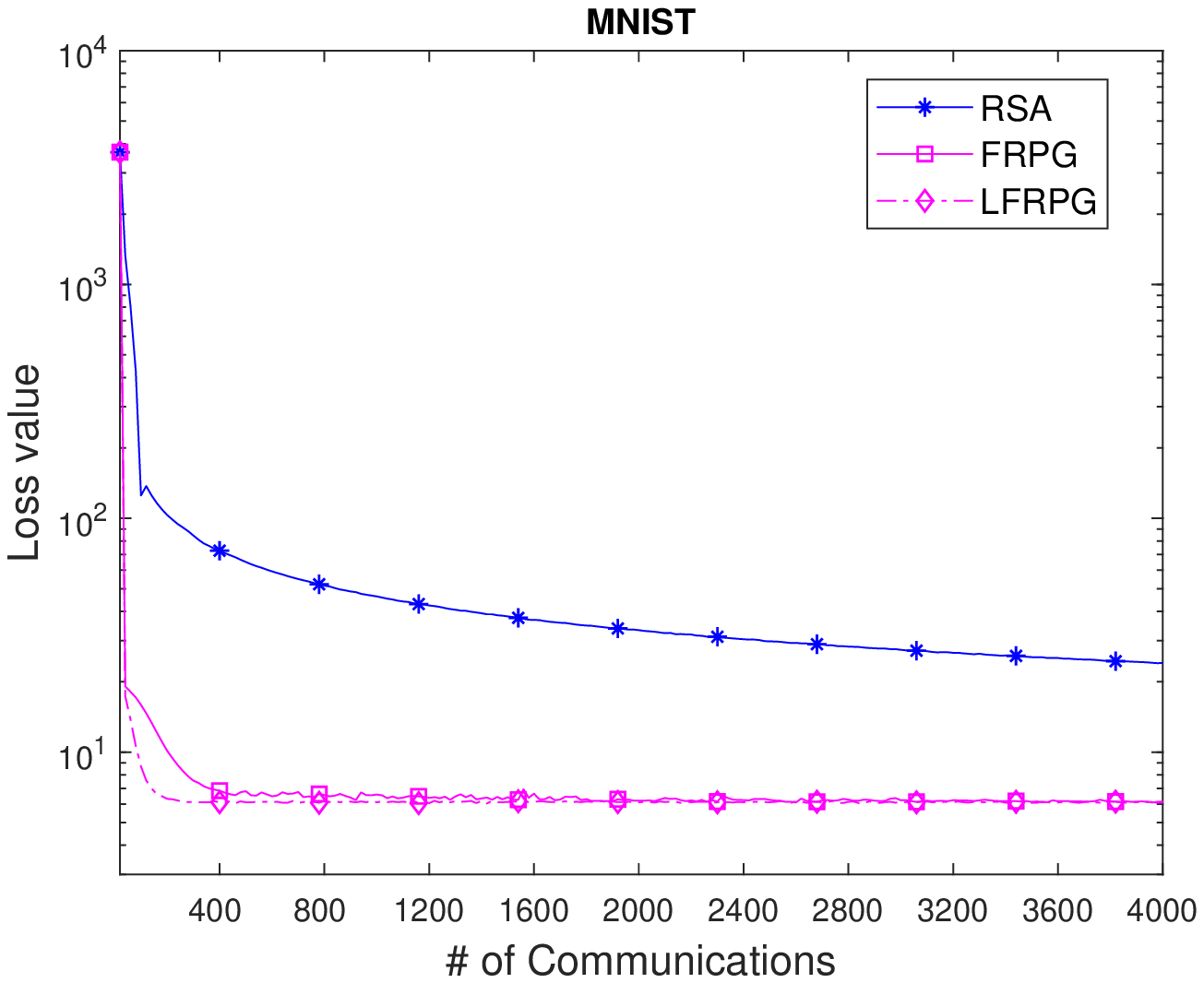}
	\end{minipage}%
\hspace{0.2 cm}
	\begin{minipage}{.3\textwidth}
		\includegraphics[width= 5.5 cm]{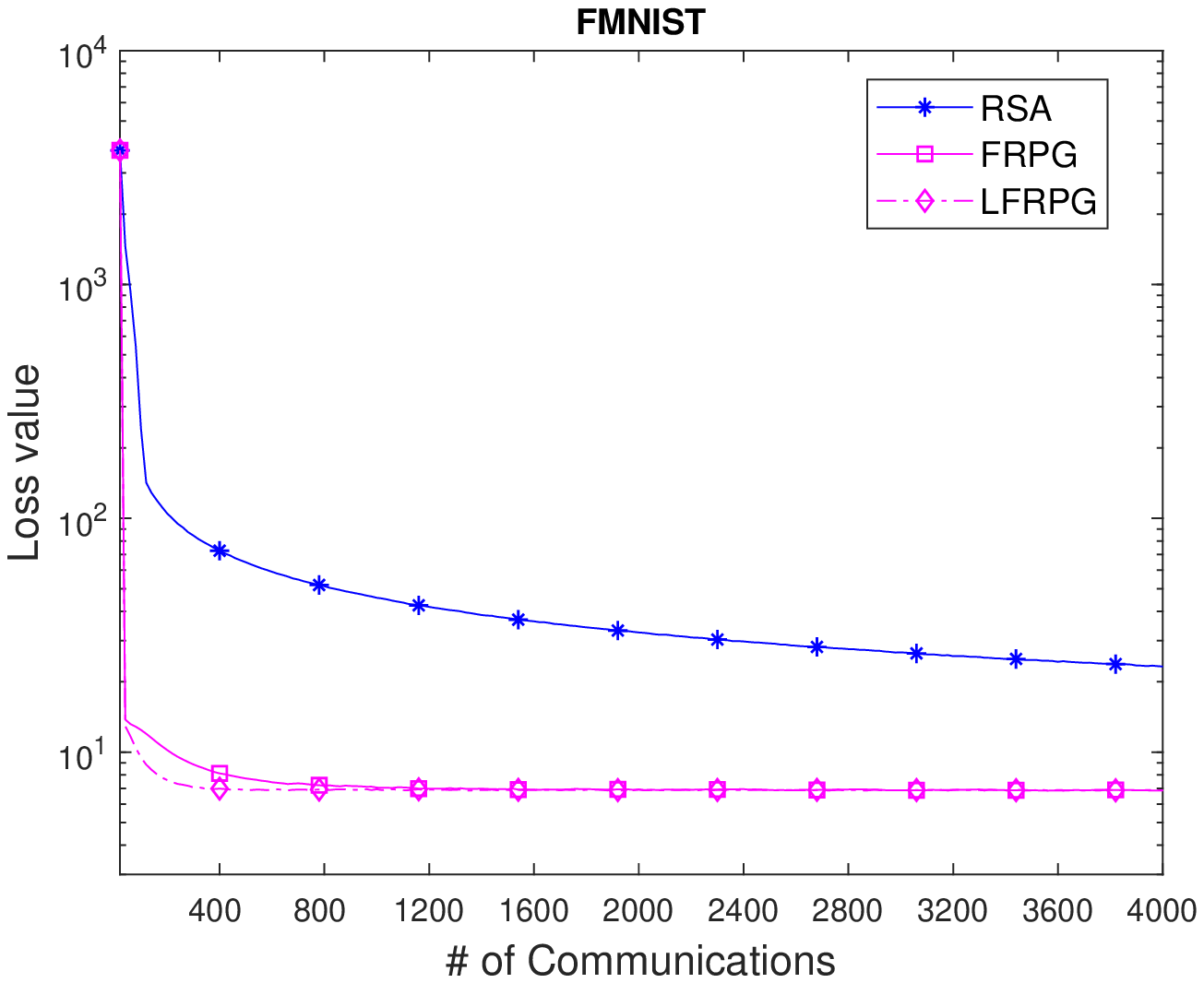}
	\end{minipage}
	\caption{The loss values over the number of communication rounds under Label-Flipping attack and heterogeneous datasets.}\label{fg:07}
	\begin{minipage}{.3\textwidth}
		\includegraphics[width= 5.5 cm]{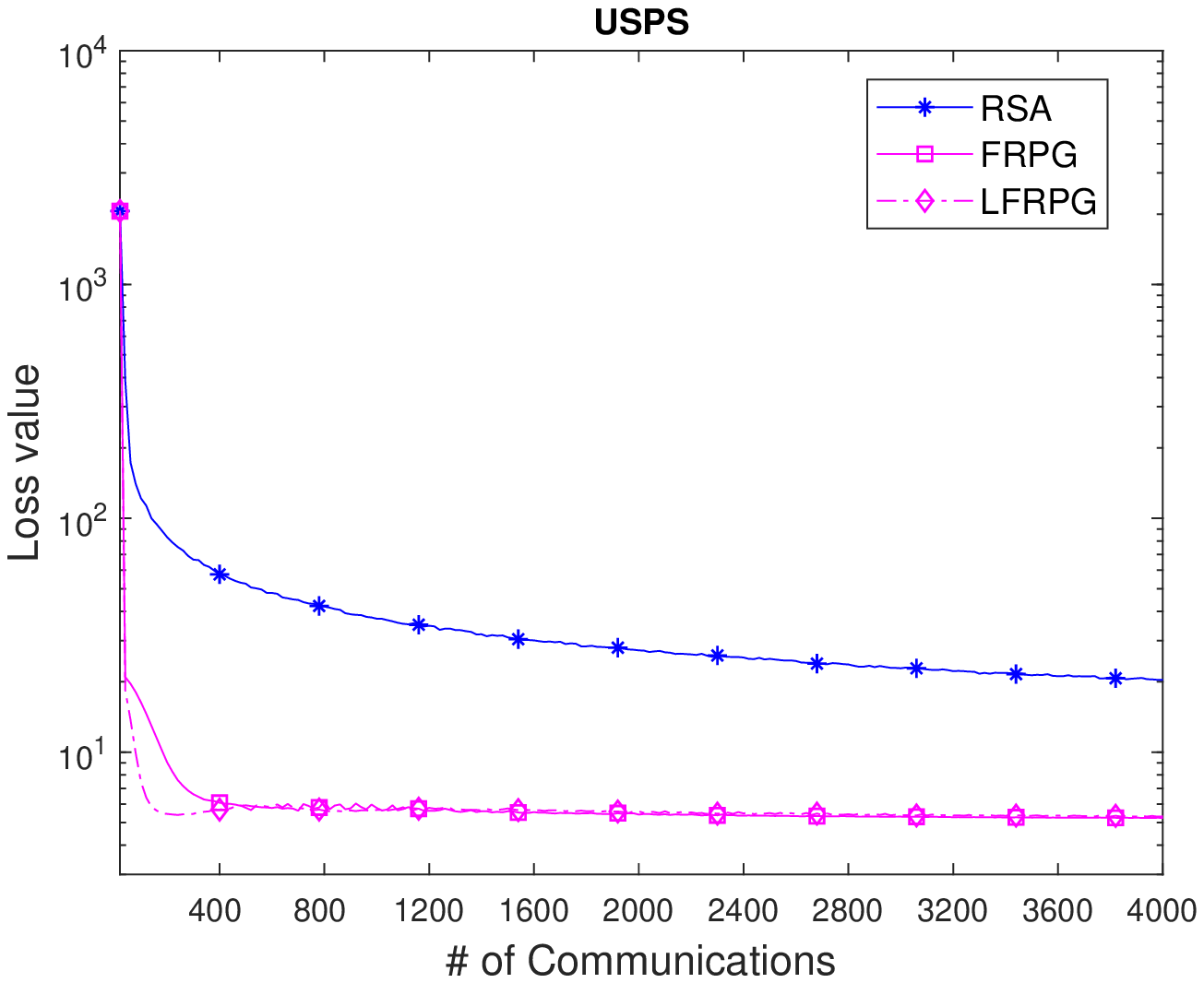}
	\end{minipage}%
\hspace{0.2 cm}
	\begin{minipage}{.3\textwidth}
		\includegraphics[width= 5.5 cm]{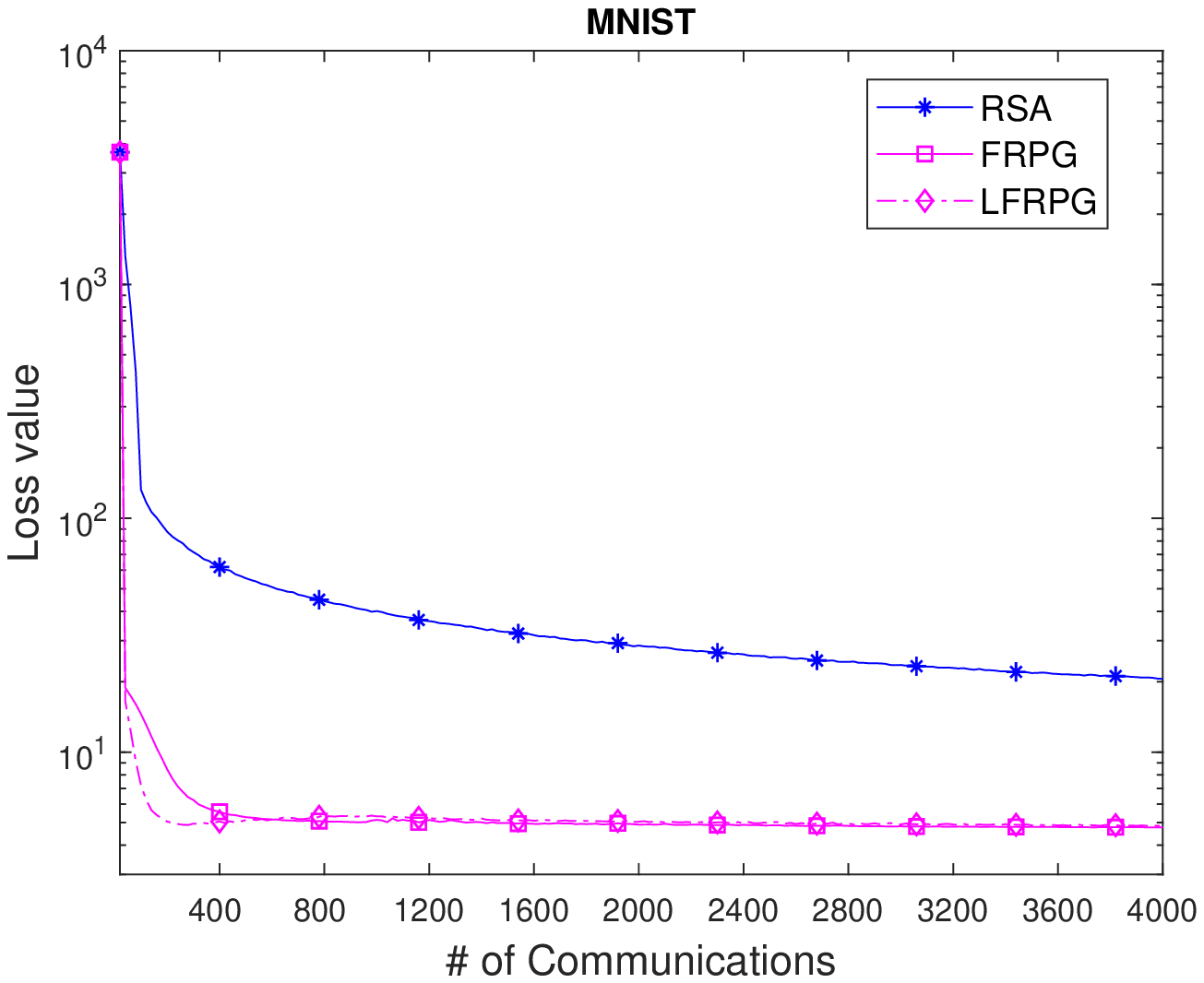}
	\end{minipage}%
\hspace{0.2 cm}
	\begin{minipage}{.3\textwidth}
		\includegraphics[width= 5.5 cm]{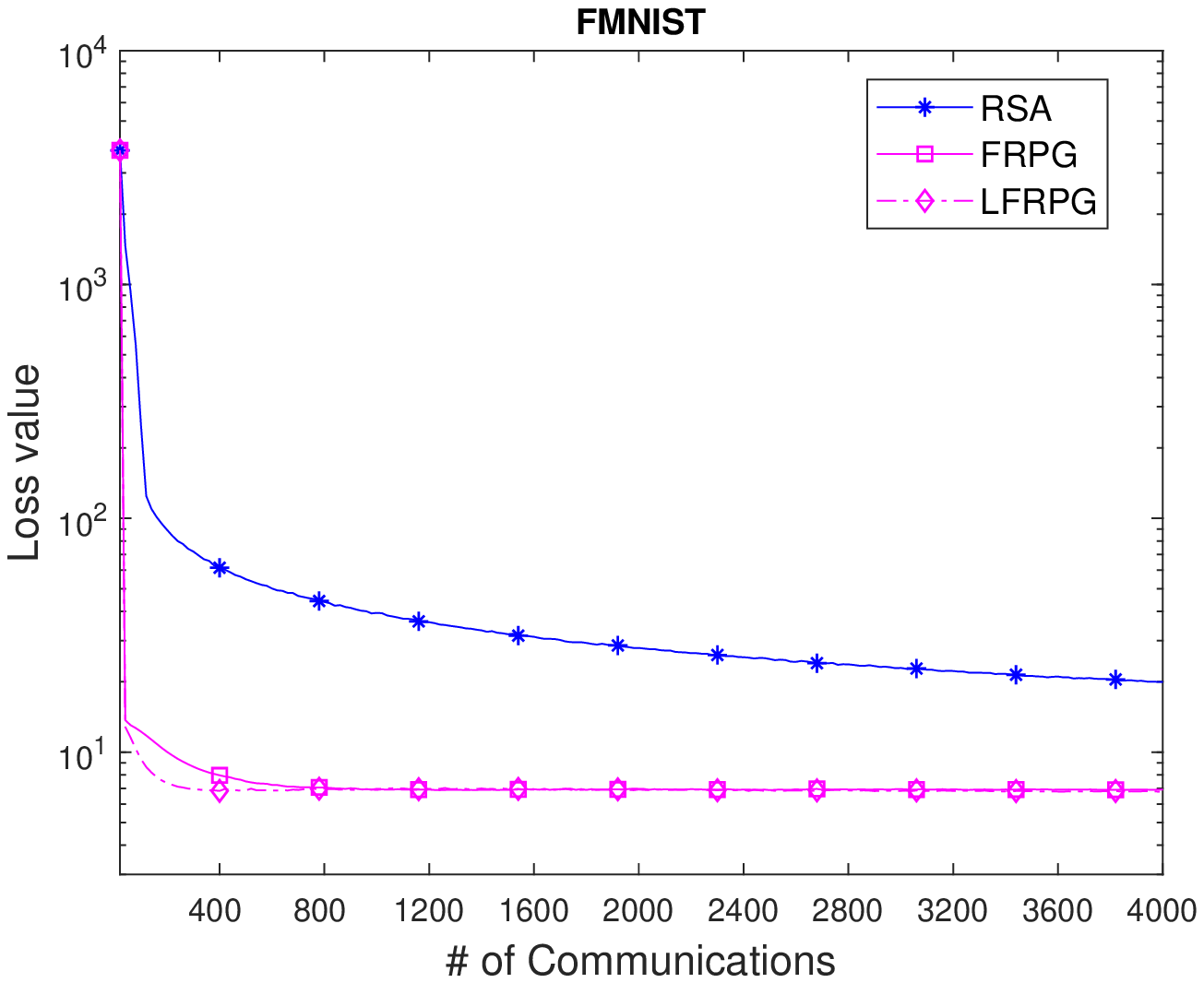}
	\end{minipage}
	\caption{The loss values over the number of communication rounds under Gaussian attack  and heterogeneous datasets.}\label{fg:08}
	\hrulefill
\end{figure*}

\section{Experiments}
To validate our analytical results, we tested the performance of FRPG and LFRPG numerically on real datasets (USPS \cite{291440}, MNIST \cite{726791} and FMNIST \cite{xiao2017fashion}).
In the USPS set, we used $8,000$ data vectors of size $256 \times 1$ for training, and $3,000$ for testing. In MNIST, we used $60,000$ data vectors of size $784 \times 1$ for training, and $10,000$ for testing. In FMNIST, we used $60,000$ data vectors of size $784 \times 1$ for training, and $10,000$ for testing. The heterogeneity of datasets was manifested as follows. 
Each pair of workers were assigned data of the same handwritten digits, and $50\%$ of the handwritten digits were removed. For example, the data samples with labels $6$, $7$, $8$ and $9$ were removed in half of the tests. We consider the Label-Flipping attack \cite{pmlr-v80-yin18a},  and the Gaussian attack \cite{Dong2019}, to verify the robustness of FRPG and LFRPG. For the Label-Flipping attack the original label $y$ was skewed to $9-y$; while for the Gaussian attack we set $w_n = c \times {\cal N}\br{0, 1}$ with $c = 1 \times 10^4$. The tests were run on  
MATLAB R2018b with Intel i7-8700 CPU @ 3.20 GHz and 16 Gb RAM. 

The multinomial logistic regression was employed as the loss with regularizer $({\delta_n}/{2})\norm{w_n}^2$. At the parameter server, we set $f_0\br{w_0} = ({\delta_0}/{2})\norm{w_0}^2$. Huber's cost with smoothing constant $\mu = 10^{-3}$ was adopted as the penalty function 
\begin{equation}
	p_n\br{w_0 - w_n} = \left\{ \begin{array}{l}
		\frac{1}{2\mu}{\norm{w_{0} - w_{n}}^2},  \norm{w_{0} - w_{n}} \le \mu \\
		\norm{w_{0} - w_{n}} - \frac{\mu}{2}, \mbox{otherwise}
	\end{array} \right .
\end{equation}
We considered a setting with $Q = 20$ workers, $N=16$ reliable ones, and weight $\lambda = 1.6$. 
The training data were evenly distributed across the workers. With faulty workers attacking by flipping labels, the mini-batch size was set to $15$; while for those adopting a Gaussian attack, the mini-batch size was set to $10$. To obtain a good top-1 accuracy convergence, we set the step sizes for benchmark schemes to $\frac{3}{\sqrt{k}}$. A strongly convex modulus with  $\delta_n = 0.003$ was chosen for $n = 0, 1, \ldots, N$; while the Lipschitz constants for the USPS, MNIST and FMNIST datasets were respectively set to $156$, $295$, and $524$. The workers in LFRPG communicated with the server every ten slots. 

We also tested communication efficiency in comparison with Krum \cite{NIPS2017_6617}, GeoMed \cite{chen2017}, and RSA \cite{Li2019} benchmarks. 
For Krum and GeoMed, the workers upload local models $\{w_{n,k}\}_{n=1}^Q$ to the servers per slot $k$. 
Then, the server performs aggregation over $\{w_{n,k}\}_{n=1}^Q$ using Krum \cite{NIPS2017_6617} and GeoMed \cite{chen2017}. Note that GeoMed for a set of vectors can be obtained by a fast Weiszfeld's algorithm \cite{weiszfeld2009point}. 
To demonstrate the negative effects of different attacks, we employed SGD by averaging heuristically the local gradients of workers. 
Figures \ref{fg:07} and \ref{fg:08} show the convergence of FRPG, LFRPG and RSA under Label-Flipping, and Gaussian attacks, respectively. After $4,000$ communication rounds, FRPG and LFRPG converge faster than RSA, while LFRPG outperforms FRPG for the same number of rounds. 
To reach the same loss value with the FMNIST dataset, LFRPG takes around $400$ communication rounds versus $800$ required by FRPG under Label-Flipping attacks. 

\begin{figure*}[t]
	\centering
	\begin{minipage}{.33\textwidth}
		\includegraphics[width= 6.1 cm]{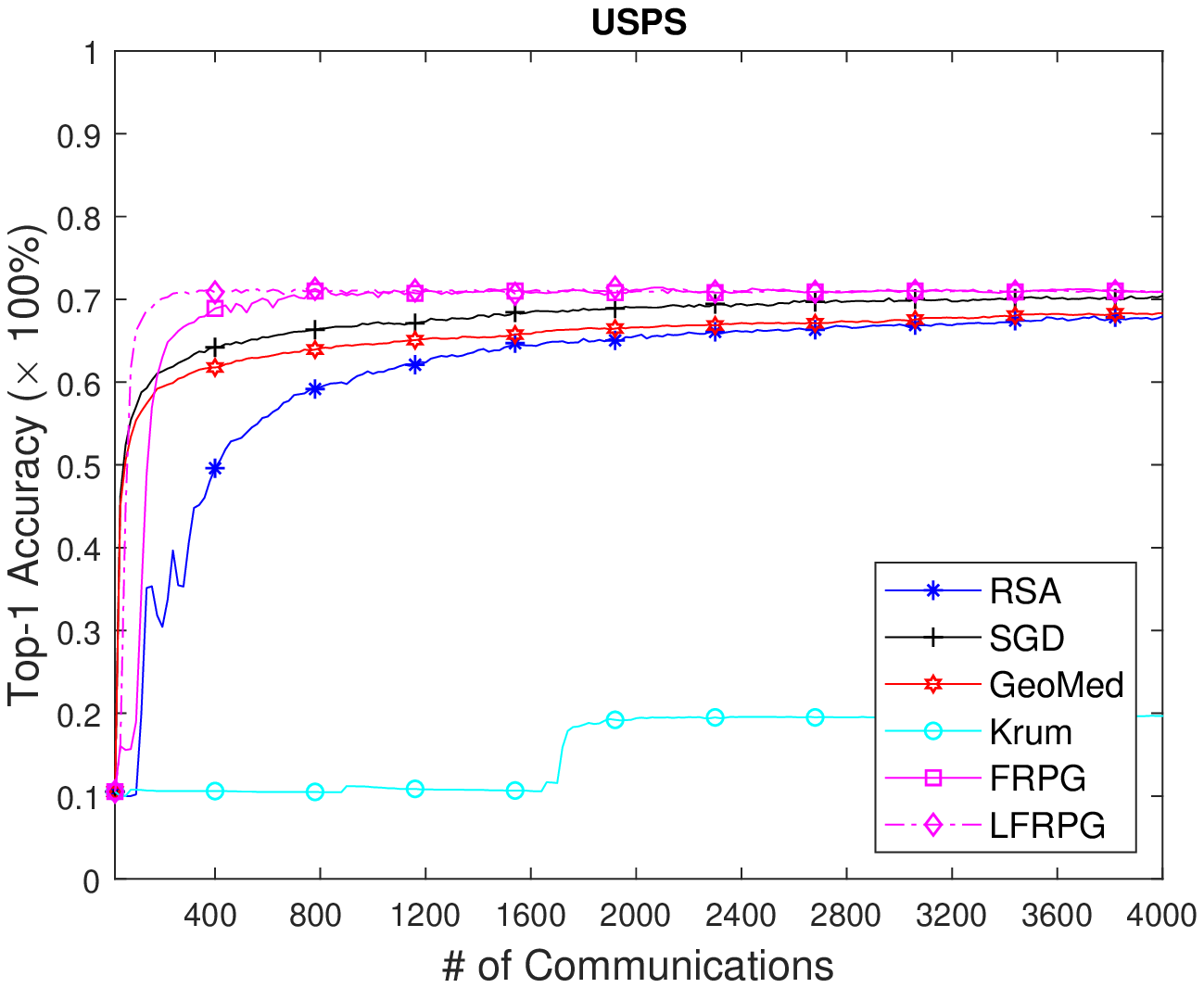}
	\end{minipage}%
	\begin{minipage}{.33\textwidth}
		\includegraphics[width= 6.1 cm]{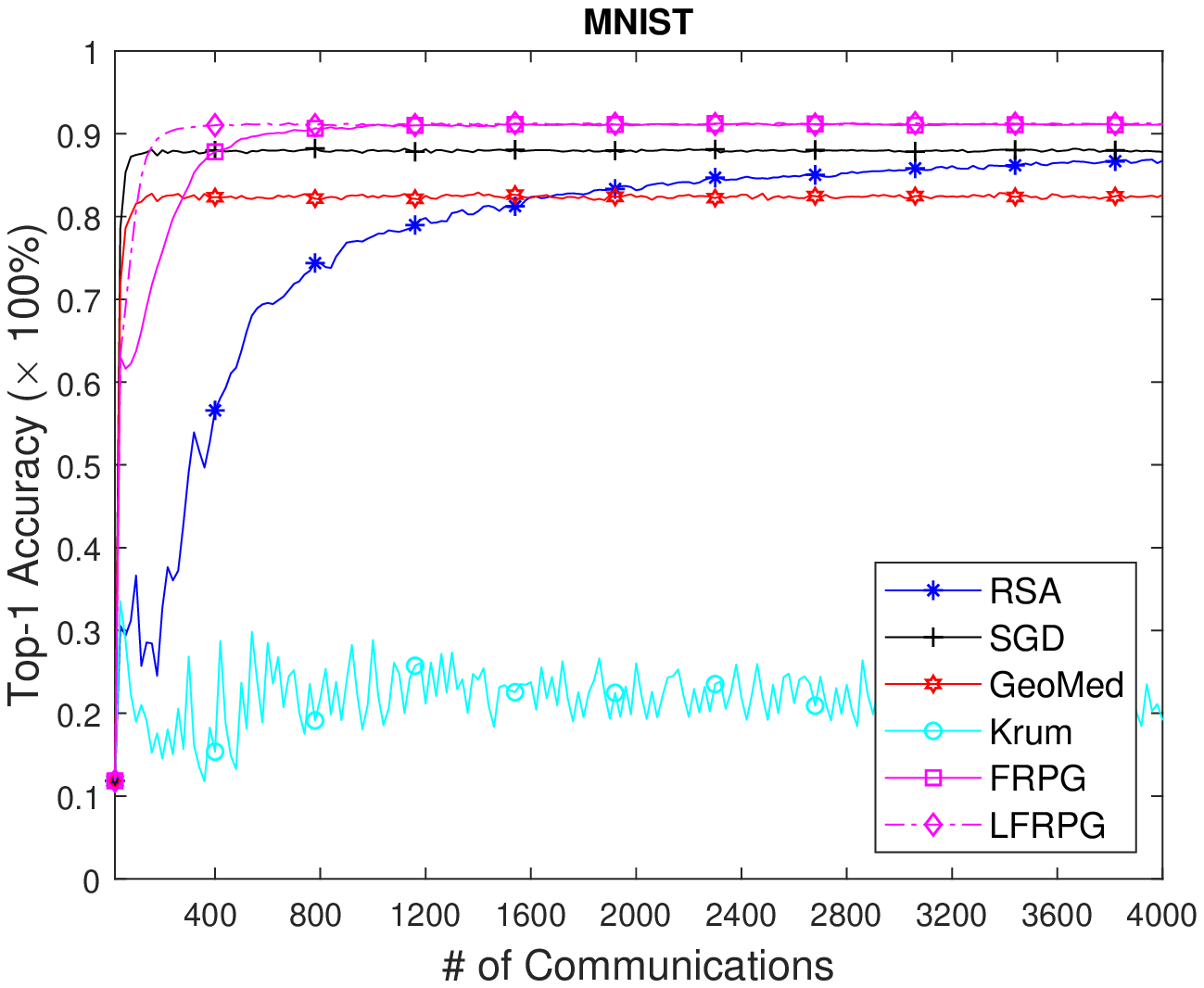}
	\end{minipage}%
	\begin{minipage}{.33\textwidth}
		\includegraphics[width= 6.1 cm]{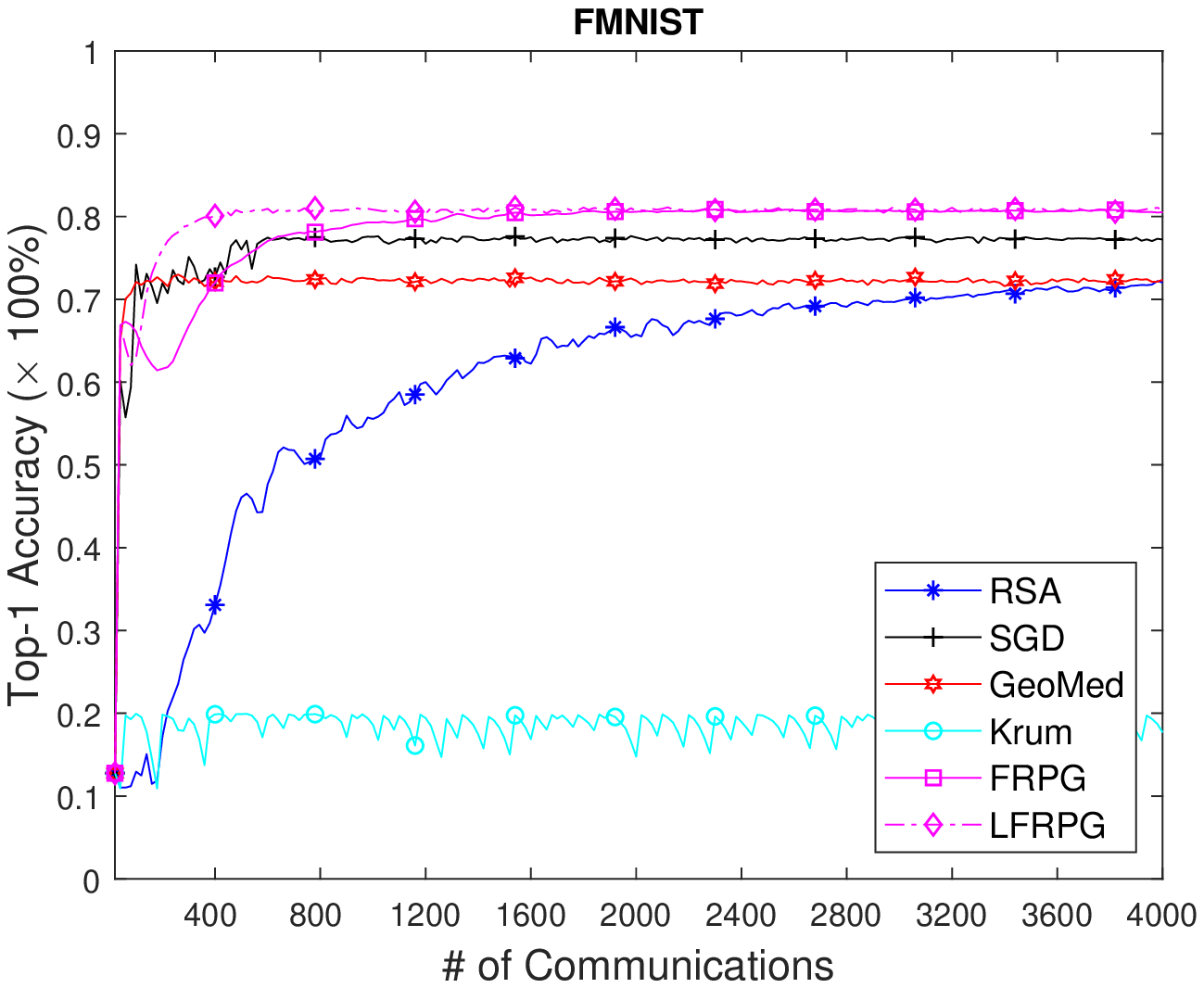}
	\end{minipage}
	\caption{Top-1 accuracy over the number of communication rounds under Label-Flipping attack  and heterogeneous datasets.}\label{fg:09}
	\begin{minipage}{.33\textwidth}
		\includegraphics[width= 6.1 cm]{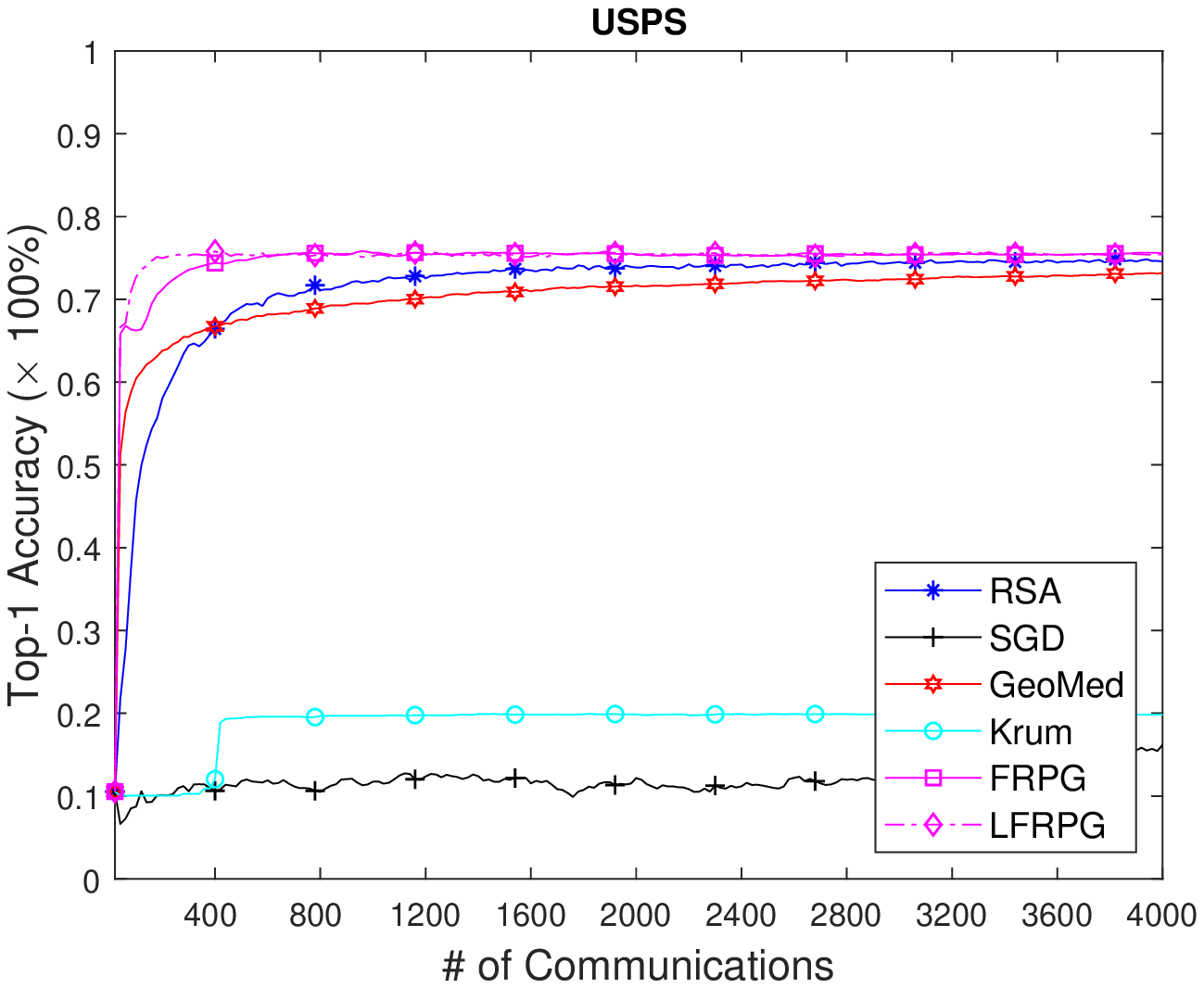}
	\end{minipage}%
	\begin{minipage}{.33\textwidth}
		\includegraphics[width= 6.1 cm]{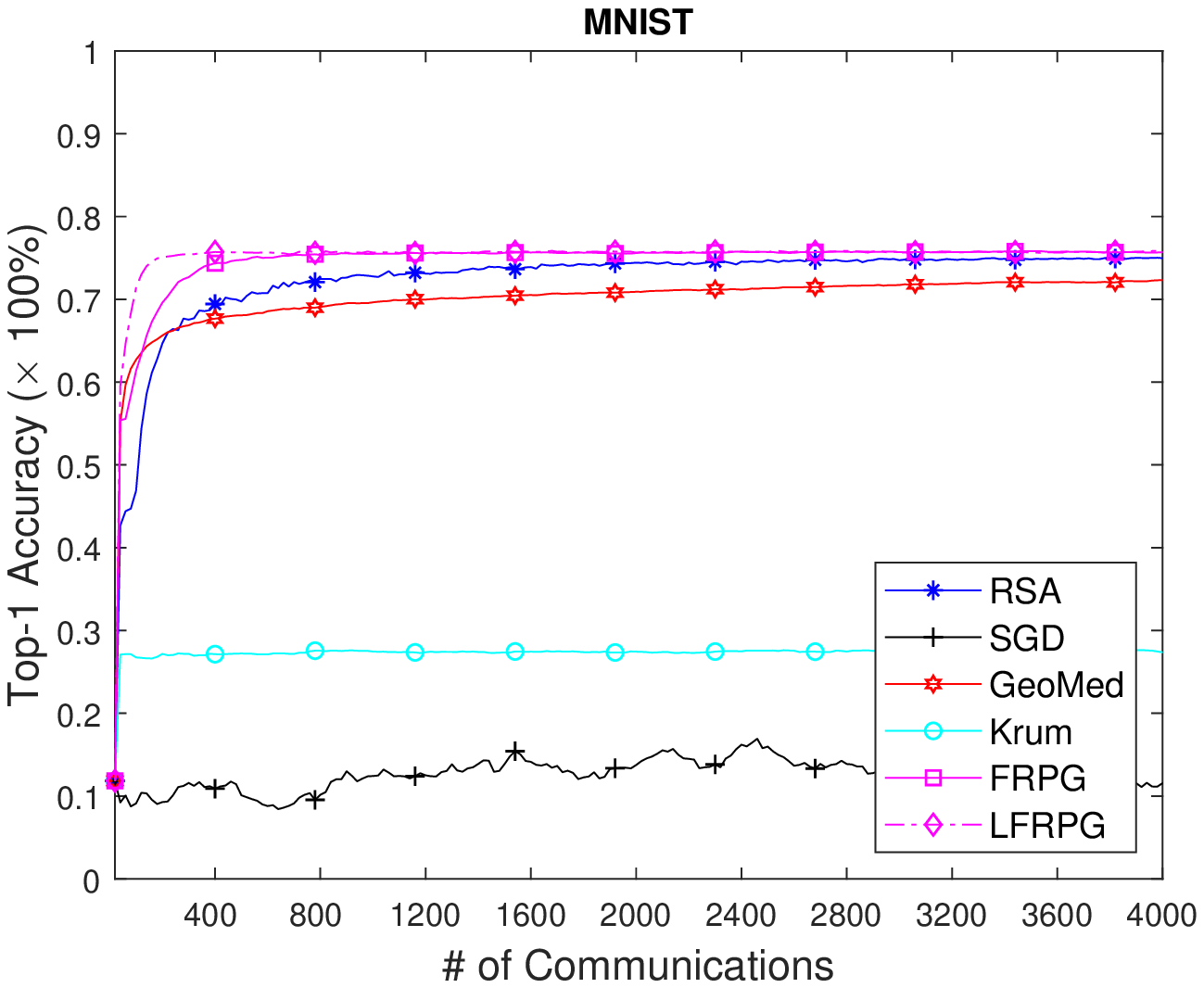}
	\end{minipage}%
	\begin{minipage}{.33\textwidth}
		\includegraphics[width= 6.1 cm]{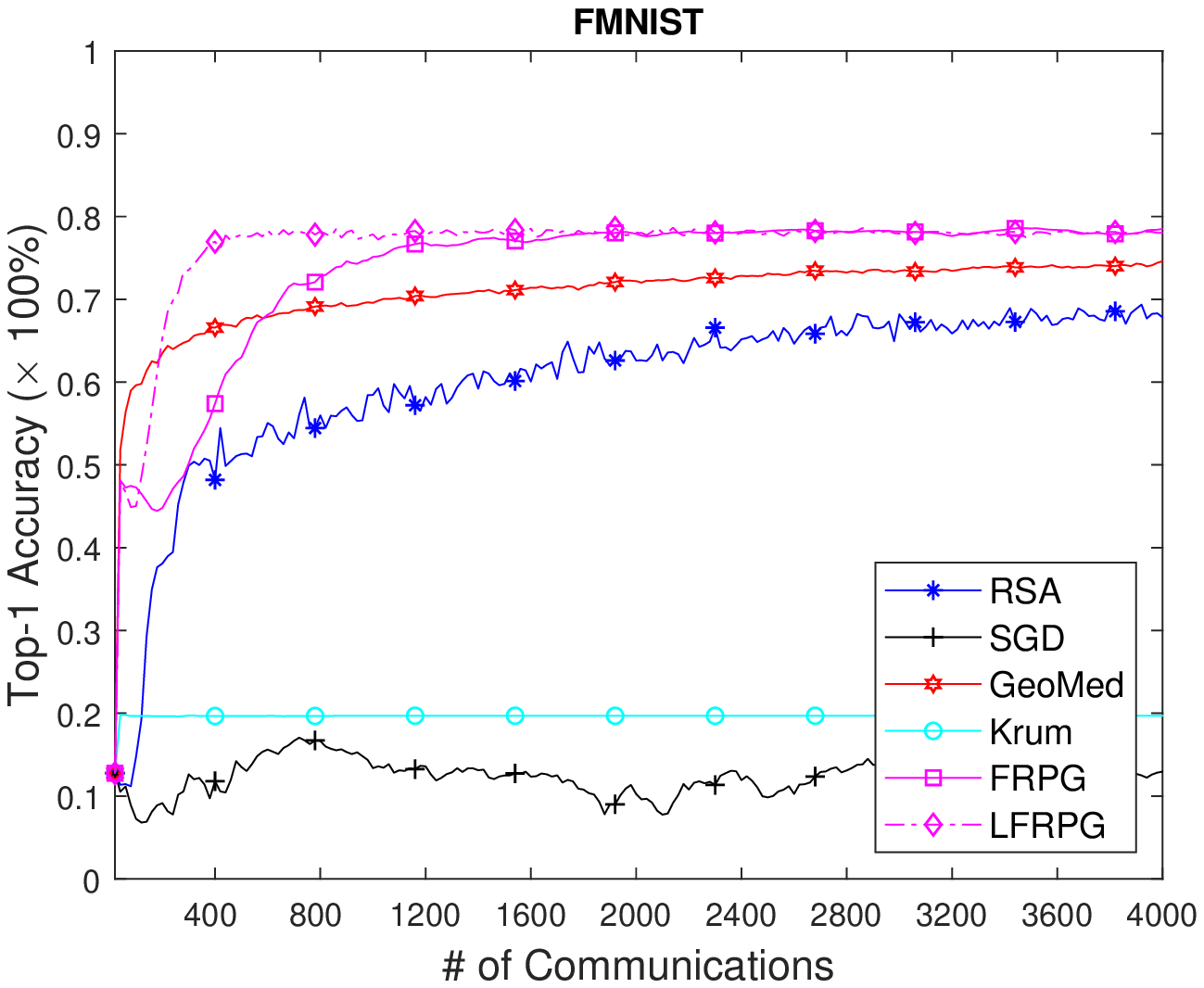}
	\end{minipage}
	\caption{Top-1 accuracy over the number of communication rounds under Gaussian attack  and heterogeneous datasets.}\label{fg:10}
	\hrulefill
\end{figure*}

Figure \ref{fg:09} compares the top-1 accuracy with Krum, GeoMed and RSA, under \mbox{Label-Flipping} attacks, respectively. Under Label-Flipping attacks, both FRPG and LFRPG converge faster than the benchmarks. FRPG and LFRPG also achieve better top-1 accuracy for the USPS, MNIST, and FMNIST datasets, while Krum fails because it is designed for homogeneous datasets. When the USPS dataset is used, we observe that the top-1 accuracy of FRPG reaches about 71\%  after about 1200 communication rounds, and the top-1 accuracy of LFRPG reaches about 71\% after 400 communication rounds. LFRPG allows agents to communicate with the parameter server every ten slots, during which each reliable agent updates the local model parameter based on the local dataset. Compared with FRPG, LFRPG has ten times the local computational cost. Therefore, we conclude that LFRPG can reduce the communication overhead at the expense of local computational cost. Besides, SGD requires 4000 communication rounds to reach the same top-1 accuracy as FRPG and LFRPG. GeoMed and RSA require more than 4000 communication rounds to achieve 71\% accuracy. In other words, FRPG and LFRPG can reduce at-least 70\% and 90\% of communication overhead when the USPS dataset is used. Besides, the reduction of communication overhead can also be observed when the MNIST and FMNIST datasets are used.

Since Label-Flipping attacks do not change the magnitude of local gradients, their negative effects on SGD are limited when heterogeneous datasets are used. For this reason, we considered the more severe Gaussian attack. Fig. \ref{fg:10} illustrates that SGD fails in the presence of Gaussian attacks. However, both FRPG and LFRPG converge faster and achieve better top-1 accuracy than Krum, GeoMed, and RSA. Using Gaussian attacks and the FMNIST dataset, the top-1 accuracy of FRPG and LFRPG is 4.13\% better than that of GeoMed, and 9.89\% better than that of RSA.

\section{Conclusions}
This work dealt with fault-resilient federated learning. Cross-fertilizing benefits of the  
robust stochastic aggregation framework and Nesterov's acceleration technique, two algorithms 
were developed to reduce the communication overhead involved. Both were proved to attain performance gains relative to the benchmarks in terms of communication efficiency. Numerical tests also confirmed this improved communication efficiency over different real datasets. 

\appendices
\section{Proof of Lemma \ref{le:01}}\label{apdx:01}
Based on the strong convexity of $f_0$, we obtain
\begin{subequations}\label{apdx1:02}
	\begin{align}
	f_0\br{u_0} \ge& f_0\br{u_{0,k}} + \inp{\nabla f_0\br{u_{0,k}}, u_0 - u_{0,k}} + \frac{\delta_0}{2}\norm{u_0 - u_{0,k}}^2 \nonumber\\
	\ge& f_0\br{w_{0,k}} - \frac{L_0}{2}\norm{w_{0,k} - u_{0,k}}^2 \nonumber\\
	& + \inp{\nabla f_0\br{u_{0,k}}, u_0 - w_{0,k}} + \frac{\delta_0}{2}\norm{u_0 - u_{0,k}}^2 \label{apdx1:02b}\\
	=& f_0\br{w_{0,k}} - \frac{L_0}{2\alpha^2_{0,k}}\norm{h_{0,k}}^2 + \frac{\delta_0}{2}\norm{u_0 - u_{0,k}}^2 \nonumber\\
	& + \inp{h_{0,k} + \sumq g_{n,k} -  \sumq g_{n,k},  u_0 -  w_{0,k}}  \label{apdx1:02c} \\
	% \ge& f_0\br{\w_{0,k}} - \frac{L_0}{2\alpha^2_{0,k}}\norm{\h_{0,k}}^2 + \frac{\delta_0}{2}\norm{\u_0 - \u_{0,k}}^2 \nonumber \\
	% & + \inp{\h_{0,k}+ \sumq\g_{n,k}, \u_0 - \u_{0,k} + \u_{0,k} - \w_{0,k}} \nonumber\\
	% & - \inp{ \sumq\g_{n,k}, \u_0 - \w_{0,k}} \label{apdx1:02d} \\
	\ge& f_0\br{w_{0,k}} + \frac{2\alpha_{0,k} - L_0}{2\alpha^2_{0,k}}\norm{h_{0,k}}^2 + \frac{\delta_0}{2}\norm{u_0 - u_{0,k}}^2 \nonumber \\
	& + \inp{h_{0,k}+ \sumq g_{n,k}, u_0 - u_{0,k}}  \nonumber\\
	& - \inp{ \sumq g_{n,k}, u_0 - w_{0,k}} - \norm{\sumq g_{n,k}}\frac{\norm{h_{0,k}}}{\alpha_{0,k}} \label{apdx1:02e}
	\end{align}
\end{subequations}
where \eqref{apdx1:02b} follows the Lipschitz continuous gradient of $f_0$; the RHS \eqref{apdx1:02c} uses the definition (cf. \eqref{eqa:brpg:01b})
$h_{0,k} := \alpha_{0,k}\br{u_{0,k} - w_{0,k}} = \nabla f_0\br{u_{0,k}}$; while the RHS of \eqref{apdx1:02e} also relies on $\langle{ \sum\nolimits_{n=1}^Q g_{n,k}, u_{0,k} - w_{0,k}}\rangle \ge -\frac{1}{\alpha_{0,k}}\|{ \sum\nolimits_{n=1}^Q g_{n,k}}\|\|{ h_{0,k}}\|$.
Substituting $h_{0,k} $ into \eqref{apdx1:02e} completes the proof. 

\section{Proof of Lemma \ref{le:02}}\label{apdx:02}
Using the proximal operator definition, rewrite \eqref{eqa:brpg:02b} as 
\begin{equation}\label{apdx2:01}
\begin{split}
w_{n,k} = &\argmin_{u_n}\left\{\inp{\nabla f\br{u_{n,k}; x_{n,k}}, u_n-u_{n,k} }  \right.\\
&\left.+ \frac{\alpha_{n,k} }{2}\norm{u_n-u_{n,k} }^2 + \lambda p_n\br{w_{0,k} - u_n}\right\}.
\end{split}
\end{equation}

Based on the update in \eqref{apdx2:01}, $\lambda\nabla_{w_n}p_n\br{w_{0,k} - w_{n,k}} = -g_{n,k}$, and the definition of gradient noise $\Delta_{n,k}$ in \eqref{eqa:brpg:03}, we obtain
\begin{equation}
h_{n,k}  :=  \alpha_{n,k}\br{u_{n,k} - w_{n,k}} = 
\nabla f\br{u_{n,k}; x_{n,k}} -  g_{n,k}  \label{apdx2:02}
\end{equation}
\begin{equation}
\nabla f_n\br{w_{n,k}} = h_{n,k} +  g_{n,k} - \Delta_{n,k} \label{apdx2:03}.
\end{equation}
Since $f_n$ is Lipschitz continuous, we deduce that 
\begin{align}
	\hspace{-0.1 cm}& \!f_n\br{u_{n,k}}\! \nonumber\\
	\hspace{-0.1 cm}	\ge & f_n\br{w_{n,k}}\! -\! \inp{\nabla f_n\br{u_{n,k}}, w_{n,k} \!-\! u_{n,k}}\!-\! \frac{L_n}{2}\norm{u_{n,k} - w_{n,k}}^2 \nonumber\\
	\hspace{-0.1 cm}	=& f_n\br{w_{n,k}} - \inp{\nabla f_n\br{u_{n,k}}, w_{n,k} - u_{n,k}}- \frac{L_n}{2\alpha_{n,k}^2}\norm{h_{n,k}}^2. \label{apdx2:04b}
\end{align}

Based on the strong convexity of $f_n$, we further obtain
\begin{equation}\label{apdx2:05}
\begin{split}
&f_n\br{u_n} \\
&\ge f_n\br{u_{n,k}} 
+ \inp{\nabla f_n\br{u_{n,k}}, u_n - u_{n,k}} + \frac{\delta_n}{2}\norm{u_n - u_{n,k}}^2\!\!.
\end{split}
\end{equation}
Summing \eqref{apdx2:04b} and \eqref{apdx2:05}, we arrive at 
\begin{subequations}\label{apdx2:06}
\begin{align}
& f_n\br{u_n} - f_n\br{w_{n,k}} \label{apdx2:05a}\\
\ge & \inp{\nabla f_n\br{u_{n,k}}, u_n - w_{n,k}} - \frac{L_n}{2\alpha_{n,k}^2}\norm{h_{n,k}}^2 \nonumber\\
& + \frac{\delta_n}{2}\norm{u_n - u_{n,k}}^2 \label{apdx2:05b} \\
\ge & \inp{h_{n,k} + g_{n,k} - \Delta_{n,k}, u_n - w_{n,k}} - \frac{L_n}{2\alpha_{n,k}^2}\norm{h_{n,k}}^2 \nonumber \\
& + \frac{\delta_n}{2}\norm{u_n - u_{n,k}}^2 \label{apdx2:05c} \\
\ge & \frac{2\alpha_{n,k} - L_n}{2\alpha_{n,k}^2}\norm{h_{n,k}}^2 + \frac{\delta_n}{2}\norm{u_n - u_{n,k}}^2 \nonumber\\
& - \inp{ \Delta_{n,k} - g_{n,k}, u_n - w_{n,k}} + \inp{h_{n,k}, u_n - u_{n,k}} \label{apdx2:05d}
\end{align}
\end{subequations}
where the RHS of \eqref{apdx2:05c} is due to \eqref{apdx2:03}, and the RHS of \eqref{apdx2:05d} follows from \eqref{apdx2:02}. 

Finally, substituting into \eqref{apdx2:05d} completes the proof. 

\section{Proof of Lemma \ref{le:03}}\label{apdx:03}
With $h_{0,k}$ as in  \eqref{apdx1:02c}, construct a strongly convex function with modulus $\delta_0 + \alpha_{0,k}\beta_k$ per slot $k$, as
\begin{align}
\phi_{0,k}\br{u_0} := &\inp{h_{0,k} +  \sumq g_{n,k}, u_0 - u_{0,k}} \label{apdx3:01}\\
& + \frac{\delta_0}{2}\norm{u_0 - u_{0,k}}^2
+ \frac{\alpha_{0,k}\beta_k}{2}\norm{u_0 - v_{0,k-1}}^2 \;.\nonumber
\end{align}
According to \eqref{eqa:brpg:01c}, $\v_0^k$ is the minimizer of \eqref{apdx3:01}. Strong convexity implies that $\phi_{0,k}\br{v_{0,k}} \le \phi_{0,k}\br{u_0} - \frac{\delta_0 + \alpha_{0,k}\beta_k}{2}\norm{v_{0,k} - u_0}^2$. Thus, upon expanding $\phi_{0,k}\br{v_{0,k}}$ and $\phi_{0,k}\br{u_0}$, we have
\begin{align}
& \inp{h_{0,k} + \sumq g_{n,k}, u_0 - u_{0,k}} + \frac{\delta_0}{2}\norm{u_0 - u_{0,k}}^2 \label{apdx3:02}\\
\ge&  \inp{h_{0,k}+ \sumq g_{n,k}, v_{0,k} - u_{0,k}}
+ \frac{\alpha_{0,k}\beta_k + \delta_0}{2}\norm{u_0 - v_{0,k}}^2  \nonumber\\
& + \frac{\alpha_{0,k}\beta_k}{2}\norm{v_{0,k} - v_{0,k-1}}^2
- \frac{\alpha_{0,k}\beta_k}{2}\norm{u_0 - v_{0,k-1}}^2. \nonumber
\end{align}

Similar to \eqref{apdx3:02}, and with $h_{n,k}$ as in \eqref{apdx2:02},
we obtain
\begin{align}
& \inp{h_{n,k}, u_n - u_{n,k}} + \frac{\delta_n}{2}\norm{u_n - u_{n,k}}^2 \label{apdx3:03}\\
\ge&  \inp{h_{n,k}, v_{n,k} - u_{n,k}}
+ \frac{\alpha_{n,k}\beta_k + \delta_n}{2}\norm{u_n - v_{n,k}}^2 \nonumber\\
& + \frac{\alpha_{n,k}\beta_k}{2}\norm{v_{n,k} - v_{n,k-1}}^2
- \frac{\alpha_{n,k}\beta_k}{2}\norm{u_n - v_{n,k-1}}^2\:. \nonumber
\end{align}

Substituting \eqref{apdx3:02} and \eqref{apdx3:03} into \eqref{eqa:brpg:07}, we thus find 
\begin{align}
& F\br{\w_k} -  F\br{\u} \label{apdx3:04}\\
\le &  \frac{\norm{\sumq g_{n,k}}}{\alpha_{0,k}}\norm{h_{0,k}}
- \sumam\frac{2\alpha_{n,k} - L_n}{2\alpha_{n,k}^2}\norm{h_{n,k}}^2 \nonumber\\
& + \sumam\frac{\eta_{1, n, k}}{\beta_k} + \sumam\inp{\Delta_{n,k}, u_n - w_{n,k}}  \nonumber\\
& + \inp{h_{0,k}+ \sumq g_{n,k}, u_{0,k} - v_{0,k}}
+ \summ\inp{h_{n,k}, u_{n,k} - v_{n,k}} \nonumber
\end{align}
where $\eta_{1, n, k}$ is defined as
\begin{align}
\eta_{1, n, k} := & \frac{\alpha_{n,k}\beta^2_k}{2}\norm{u_n - v_{n, k-1}}^2 \nonumber\\
& - \frac{\delta_n\beta_k + \alpha_{n,k}\beta^2_k}{2}\norm{u_n - v_{n,k}}^2 \nonumber\\
& - \frac{\alpha_{n,k}\beta^2_k}{2}\norm{v_{n,k} - v_{n,k-1}}^2. \label{apdx3:05}
\end{align}
Setting $\u = \w_{k-1}$ in \eqref{eqa:brpg:07}, and dropping the non-positive terms $-\frac{\delta_n}{2}\|{w_{n,k-1} - w_{n,k}}\|^2$, we arrive at 
\begin{equation}\label{apdx3:06}
	\begin{split}
	& F\br{\w_k} -  F\br{\w_{k-1}}  \\
	\le & \frac{\norm{\sumq g_{n,k}}}{\alpha_{0,k}}\norm{h_{0,k}}
	- \sumam\frac{2\alpha_{n,k} - L_n}{2\alpha_{n,k}^2}\norm{h_{n,k}}^2  \\
	& + \sumam\inp{\Delta_{n,k}, w_{n,k-1} - w_{n,k}} \\
	& + \inp{h_{0,k} +  \sumq g_{n,k}, u_{0,k} - w_{0,k-1}} \\
	& + \summ\inp{h_{n,k}, u_{n,k} - w_{n,k-1}}. 
	\end{split}
\end{equation}
Using \eqref{apdx3:04} and \eqref{apdx3:06}, the convex combination $\beta_k\br{F\br{\w_k} -  F\br{\u} } + \br{1-\beta_k}\br{F\br{\w_k} -  F\br{\w_{k-1}}}$ is bounded as
\begin{equation}\label{apdx3:07}
\begin{split}
& F\br{\w_k} -  F\br{\u} - \br{1-\beta_k}\br{F\br{\w_{k-1}} -  F\br{\u}} \\
\le & \frac{\norm{\sumq g_{n,k}}}{\alpha_{0,k}}\norm{h_{0,k}}
- \sumam\frac{2\alpha_{n,k} - L_n}{2\alpha_{n,k}^2}\norm{h_{n,k}}^2 \\
& + \sumam\eta_{1, n, k}  + \sumam\eta_{2, n, k}  + \sumam\eta_{3, n, k}  
\end{split}
\end{equation}
where 
\begin{equation}\label{apdx3:08} 
 \eta_{2, n, k}  :=  \inp{\Delta_{n,k}, \beta_k u_n + \br{1-\beta_k}w_{n,k-1} - w_{n,k}} 
\end{equation}
and 
\begin{align}\label{apdx3:09}
& \eta_{3, n, k}  \\
:= & \left\{ \begin{array}{l}
\inp{h_{0,k} +  \sumq g_{n,k}, u_{0,k} - \beta_k v_{0,k} - \br{1-\beta_k} w_{0,k-1}}, n= 0 \\
\inp{h_{n,k}, u_{n,k} - \beta_k v_{n,k} - \br{1-\beta_k} w_{n,k-1}}, n= 1, \ldots, N. 
\end{array}
\right. \nonumber
\end{align}

Based on \eqref{eqa:brpg:02a}, we obtain
\begin{equation}\label{apdx3:11}
\br{1-\beta_k}w_{n,k} = u_{n,k} - \beta_k v_{n,k}, n = 0, 1, \ldots, N.
\end{equation}
Substituting \eqref{apdx3:11} into \eqref{apdx3:08}, it holds for $n = 0, 1,\ldots, N$ that 
\begin{align}
\eta_{2, n, k} = & \beta_k\inp{\Delta_{n,k}, u_n - v_{n,k-1}} + \inp{\Delta_{n,k}, u_{n,k} - w_{n,k}} \nonumber\\
\le & \beta_k\inp{\Delta_{n,k}, u_n - v_{n,k-1}} + \frac{\sqrt{\norm{\Delta_{n,k}}^2}}{\alpha_{n,k}}\norm{h_{n,k}} \label{apdx3:12}
\end{align}
where \eqref{apdx3:12} follows from H\"{o}lder's inequality~\cite{Beckenbach2012}.

Taking expectation on both sides of \eqref{apdx3:12} for terms $n = 1, \ldots, N$, we obtain
\begin{align}
& \E_{x_{n, 1:K}}\sq{ \eta_{2, n, k} } \label{apdx3:13}\\
\le& \E_{x_{n, 1:K}}\sq{\beta_k\inp{\Delta_{n,k} u_n - v_{n,k-1}} + \frac{\sqrt{\norm{\Delta_{n,k}}^2}}{\alpha_{n,k}}\norm{h_{n,k}}}  \nonumber \\
=& \frac{\sigma_n}{\alpha_{n,k}}\norm{h_{n,k}} \nonumber
\end{align}
where the equality is due to the facts
\begin{equation}\label{apdx3:14}
\begin{split}
 \E_{\x_{n, 1:K}}\sq{\sqrt{\norm{\Delta_{n,k}}^2}} 
\le & \E_{\x_{n, 1:K-1}}\sqrt{\E_{x_{n,K}}\sq{\norm{\Delta_{n,k}}^2}} = \sigma_n
\end{split}
\end{equation}
and since Assumption \ref{as:04} dictates $\E_{x_{n, K}}[{\Delta_{n,k}}] = 0$, we have
\begin{equation}\label{apdx3:15}
\begin{split}
& \E_{x_{n, 1:K}}\sq{\inp{\Delta_{n,k}, u_n - v_n^{k-1}}} \\
=& \E_{x_{n, 1:K-1}}\inp{\E_{x_{n, K}}\sq{\Delta_{n,k}}, u_n - v_{n,k-1}} = 0\:.
\end{split}
\end{equation}
Based on the Young's inequality \cite{Beckenbach2012}, $\eta_{2, 0, k}$ is bounded as \begin{equation}\label{apdx3:16}
\eta_{2, 0, k}  \le \frac{\beta_k}{2\epsilon}\norm{\Delta_{0,k}}^2
+ \frac{\epsilon\beta_k}{2}\norm{u_0 - v_{0,k-1}}^2 
+ \frac{\norm{\Delta_{0,k}}}{\alpha_{0,k}}\norm{h_{0,k}}
\end{equation}
with $\epsilon \in \br{0, \infty}$.

Substituting \eqref{apdx3:11} into \eqref{apdx3:09}, we obtain for $n = 1, \ldots, N$ that 
\begin{equation}\label{apdx3:17}
\begin{split}
\eta_{3, n, k} =& \beta_k\inp{h_{n,k}, v_{n,k-1} - v_{n,k}} \\
\le& \frac{1}{2\alpha_{n,k}}\norm{h_{n,k}}^2 + \frac{\alpha_{n,k}\beta_k^2}{2}\norm{v_{n,k-1} - v_{n,k}}^2
\end{split}
\end{equation}
where the inequality is due to Young's inequality~\cite{Beckenbach2012}.

Substituting \eqref{apdx3:11} into $\eta_{3, 0, k}$, we deduce
\begin{align}
\eta_{3, 0, k}  =& \beta_k\inp{h_{0,k} +  \sumq g_{n,k}, v_{0,k-1} - v_{0,k}} \label{apdx3:18}\\
\le& \frac{1}{2\alpha_{0,k}}\norm{h_{0,k} + \sumq g_{n,k}}^2 + \frac{\alpha_{0,k}\beta_k^2}{2}\norm{v_{0,k-1} - v_{0,k}}^2  \nonumber\\
\le& \frac{2}{3\alpha_{0,k}}\norm{h_{0,k}}^2
+ \frac{2}{\alpha_{0,k}}\norm{\sumq g_{n,k}}^2  + \frac{\alpha_{0,k}\beta_k^2}{2}\norm{v_{0,k-1} - v_{0,k}}^2 \nonumber
\end{align}
where first inequality is due to Young's inequality \cite{Beckenbach2012}, and the second inequality is based on the fact that 
\begin{equation}\label{apdx3:19}
\norm{h_{0,k} + \sumq g_{n,k}}^2 \le \frac{4}{3}\norm{h_{0,k}}^2 + 4\norm{\sumq g_{n,k}}^2. 
\end{equation}
Substituting \eqref{apdx3:05}, \eqref{apdx3:12} and \eqref{apdx3:16}--\eqref{apdx3:18} into the RHS of \eqref{apdx3:07}, we obtain
\begin{align}
& F\br{\w_k} -  F\br{\u} - \br{1-\beta_k}\br{F\br{\w_{k-1}} -  F\br{\u}} \label{apdx3:20}\\
\le& \sumam \br{ \eta_{4, n, k} + \eta_{5, n, k} }
+ \frac{2}{\alpha_{0,k}}\norm{\sumq g_{n,k}}^2
+ \frac{\beta_k}{2\epsilon}\norm{\Delta_{0,k}}^2 \nonumber
\end{align}
where 
\begin{align}
&\eta_{4, n, k}  \label{apdx3:21}\\
:= &  \left\{ \begin{array}{l}
\frac{\norm{\sumq g_{n,k}} + \norm{\Delta_{0,k}}}{\alpha_{0,k}}\norm{h_{0,k}}
- \frac{2\alpha_{0,k} - 3L_0}{6\alpha_{0,k}^2}\norm{h_{0,k}}^2, n=0\\
\frac{\sigma_n}{\alpha_{n,k}}\norm{h_{n,k}} - \frac{\alpha_{n,k} - L_n}{2\alpha_{n,k}^2}\norm{h_{n,k}}^2, n = 1, \ldots, N
\end{array} \right. \nonumber
\end{align}
and 
\begin{align}
\eta_{5, n, k} 
:=  \left\{ \begin{array}{l}
\frac{\epsilon\beta_k+ \alpha_{0,k}\beta_k^2}{2}\norm{u_0 - v_{0,k-1}}^2 \\
\hspace{0.2 cm} - \frac{\delta_0\beta_k + \alpha_{0,k}\beta_k^2}{2}\norm{u_0 - v_{0,k}}^2, n = 0 \\
\frac{\alpha_{n,k}\beta_k^2}{2}\norm{u_n - v_{n,k-1}}^2 \\
\hspace{0.2 cm} - \frac{\delta_n\beta_k + \alpha_{n,k}\beta_k^2}{2}\norm{u_n - v_{n,k}}^2, n = 1, \ldots, N.
\end{array} \right.  \label{apdx3:22}
\end{align}

Using the inequality $-ax^2 + bx \le \frac{b^2}{4a}$ and the power of $\|{\sum\nolimits_{n=1}^Q g_{n,k}}\| + \|{\Delta_{0,k}}\|$ in \eqref{eqa:brpg:08}, we can bound $\eta_{4, n, k}$ as 
\begin{equation}\label{apdx3:23}
\eta_{4, n, k} \le \eta_{6, n, k} = \left\{ \begin{array}{l}
\frac{3\sigma_0^2}{2\br{2\alpha_{0,k} - 3 L_0 }}, n = 0\\
\frac{\sigma_n^2}{2\br{\alpha_{n,k} - L_n }}, n = 1, \ldots, N.
\end{array} \right.
\end{equation}

Substituting \eqref{apdx3:23} into \eqref{apdx3:20} and setting $\u = \u^*$ lead to \eqref{eqa:brpg:09}.

\section{Proof of Theorem \ref{th:01}}\label{apdx:04}
Dividing both sides of \eqref{eqa:brpg:09} by $\beta_k^2$, we can write
\begin{equation}\label{apdx4:01}
\begin{split}
& \frac{1}{\beta_k^2} \br{F\br{\w_k} - F\br{\u^*}} \\
\le & \frac{1-\beta_k}{\beta_k^2}\br{F\br{\w_{k-1}} - F\br{\u^*}} + \sumam \frac{ \eta_{5, n, k} + \eta_{6, n, k} }{\beta_k^2} \\
& + \frac{2 \lambda^2 Q^2 G}{\alpha_{0,k}\beta_k^2}
+ \frac{\lambda^2 B^2 G}{2\epsilon\beta_k}.
\end{split}
\end{equation}

Setting $\beta_k = \frac{2}{k+2}$, we can readily verify that 
\begin{equation}\label{apdx4:02}
\frac{1-\beta_k}{\beta_k^2} \le \frac{1}{\beta_{k-1}^2}.
\end{equation}

Summing \eqref{apdx4:01} over $k = 1, \ldots, K$, it follows after straightforward manipulations that
\begin{equation}\label{apdx4:03}
\begin{split}
& \frac{1}{\beta_k^2} \br{F\br{\w_k} - F\br{\u^*}} \\
\le & {F\br{\w_{0}} - F\br{\u^*}}
+ \sumk {\frac{{{\lambda ^2}{B^2} G}}{{2\epsilon \beta _k^{}}}} \\
&
+ \frac{{\frac{\epsilon}{\beta _1}  + {\alpha _{0,1}}}}{2}{\left\| {u_0^* - v_{0,0}} \right\|^2}
+ \summ{\frac{{{\alpha _{n,1}}}}{2}{{\left\| {u_n^* - v_{n,0}} \right\|}^2}}\\
& + \sum\limits_{k=1}^{K-1}\sumam \eta_{7, n, k} + \sumk\sumam \eta_{8, n, k}
\end{split}
\end{equation}
where 
\begin{align}
& \eta_{7, n, k} \label{apdx4:04}\\
:= & \left\{\begin{array}{l}
\frac{1}{2} {\left( {{\alpha _{0,k + 1}} - {\alpha _{0,k}} + \frac{\epsilon}{\beta _{k + 1}}
- \frac{\delta _0}{\beta _k}} \right){{\left\| {u_0^* - v_{0,k}} \right\|}^2}}, n = 0 \\
\frac{1}{2} {\left( {{\alpha _{n,k + 1}} - {\alpha _{n,k}} - \frac{\delta _n}{\beta _k}} \right){{\left\| {u_n^* - v_{n,k}} \right\|}^2}}, n = 1,\ldots, N
\end{array}\right. \nonumber
\end{align}
and 
\begin{equation}\label{apdx4:05}
\eta_{8, n, k}
:=  \left\{\begin{array}{l}
\frac{8\lambda^2Q^2 G + 3\sigma_0^2 }{2\br{2\alpha_{0,k} - 3 L_0}\beta_k^2}, n = 0 \\
{\frac{{\sigma _n^2}}{{2 \br{{\alpha _{n,k}} - {L_n}} \beta _k^2}}}, n = 1, \ldots, N.
\end{array}\right.
\end{equation}

To analyze the convergence of FRPG, we introduce the following constraints
\begin{subequations}\label{apdx4:06}
\begin{align}
\frac{\delta_0}{\beta_k} - \frac{\epsilon}{\beta_{k+1}} &> 0 \label{apdx4:06a}\\
\frac{\delta_0}{\beta_k} - \frac{\epsilon}{\beta_{k+1}} &\ge \alpha_{0, k+1} - \alpha_{0, k} \label{apdx4:06b}\\
\alpha_{0, k} &= \frac{3}{2}\br{\frac{c_0}{\beta_k^2} + L_0} \label{apdx4:06c}\\
\frac{\delta_n}{\beta_k} &\ge \alpha_{n, k+1} - \alpha_{n, k}, n = 1, \ldots, N \label{apdx4:06d}\\
\alpha_{n, k} &= \frac{c_n}{\beta_k^2} + L_n, n = 1, \ldots, N \label{apdx4:06e}
\end{align}
\end{subequations}
where $c_n > 0$ with $n = 0, 1, \ldots, N$; and \eqref{apdx4:06a} with $\beta_k = \frac{2}{k+2}$ imply that  $\epsilon < \frac{3}{4}\delta_0$. Without loss of generality, we set $\epsilon = \frac{1}{2}\delta_0$. Based on \eqref{apdx4:06b} and \eqref{apdx4:06c}, we have $c_0 \le \frac{4}{21}\delta_0$. Hence, $\alpha_{0,k}$ is given by $\alpha_{0,k} = \frac{\delta_0}{14}\br{k+2}^2 + \frac{3}{2}L_0$. From \eqref{apdx4:06d} and \eqref{apdx4:06e}, we deduce that $c_n \le \frac{6}{7}\delta_n$, which implies that $\alpha_{n,k} = \frac{3\delta_n}{14}\br{k+2}^2 + L_n$ with $n = 1, \ldots, N$. As a result, we find
\begin{equation}\label{apdx4:07}
\alpha_{n,k} =
\left\{\begin{array}{l}
\frac{\delta_0}{14}\br{k+2}^2 + \frac{3}{2}L_0, n = 0\\
\frac{3\delta_n}{14}\br{k+2}^2 + L_n, n = 1, \ldots, N.
\end{array}\right.
\end{equation}

Based on \eqref{apdx4:07} and $\epsilon = \frac{1}{2}\delta_0$, we simplify $\sum\nolimits_{k=1}^{K-1}\eta_{7, n, k}$ as  
\begin{equation}\label{apdx4:08}
\sum\limits_{k=1}^{K-1}\eta_{7, n, k} \le \eta_{9, n}
:= \left\{\begin{array}{l}
\br{\frac{3}{8}\delta_0 + \frac{1}{2}\alpha_{0,1}}\norm{u_0^* - v_{0,0}}^2, n = 0 \\
\frac{1}{2}\alpha_{n,1}\norm{u_n^* - v_{n,0}}^2, n = 1, \ldots, N.
\end{array}\right.
\end{equation}

Using \eqref{apdx4:07}, $\eta_{8, n, k}$ reduces to  
\begin{equation}\label{apdx4:09}
\eta_{8, n, k} = \eta_{10, n}
:= \left\{\begin{array}{l}
\frac{7\lambda^2Q^2 G + \frac{21}{8}\sigma_0^2}{\delta_0}, n = 0 \\
\frac{7\sigma^2_n}{12\delta_n}, n = 1, \ldots, N.
\end{array}\right.
\end{equation}

Substituting $\beta_k = \frac{2}{k+2}$ and \eqref{apdx4:07}--\eqref{apdx4:09} into \eqref{apdx4:03}, we establish the convergence rate of FRPG as
\begin{equation}\label{apdx4:11}
\begin{split}
 F\br{\w_k} - F\br{\u^*}  
\le & \frac{4}{\br{K+2}^2}\br{F\br{\w_0} - F\br{\u^*}+ \sumam \eta_{9, n} } \\
& + \frac{4 K }{\br{K+2}^2} \sumam\eta_{10, n} + \OO{\frac{\lambda^2B^2 G}{\delta_0}}
\end{split}
\end{equation}
where $\OO{x}$ represents a polynomial of $x$.

\section{Proof of Theorem \ref{th:02}}\label{apdx:06}
Setting $\beta^i = \frac{2}{i+2}$ so that
$\frac{1-\beta^i}{\br{\beta^i}^2} \le \frac{1}{\br{\beta^{i-1}}^2}$; summing  \eqref{eqa:lbrpg:06} over $i = 1, \ldots, I$; and, multiplying by $\br{\beta^I}^2$, we obtain \eqref{apdx6:01}. 
\begin{figure*}
\begin{equation}\label{apdx6:01}
\begin{split}
 \frac{1}{T}\sumkt F\br{\w_k^I} - F\br{\u^*}  
\le &
\br{\beta^I}^2\br{\frac{1}{T}\sumkt F\br{\w_k^{0}} - F\br{\u^*}}
+ \br{\beta^I}^2\sumam\sumi \eta_{14, n}^i + \br{\beta^I}^2\sumi \frac{\lambda^2B^2 G  }{2\epsilon\beta^i} \\
&  + \br{\beta^I}^2\br{\frac{\epsilon + \alpha_0^1\beta^1}{2\beta^1}\norm{\u_0^* - \v_0^0}^2
	+ \frac{1}{2}\sum\limits_{i=1}^{I-1}\br{\alpha_0^{i+1} - \alpha_0^{i}
		+ \frac{\epsilon}{\beta^{i+1}} - \frac{\delta_0}{\beta^{i}}}\norm{\u_0^* - \v_0^i}^2 } \\
&+ \frac{\br{\beta^I}^2}{T}\summ\br{\frac{\alpha_n^1}{2}\norm{\u_n^* - \v_n^0}^2
	+ \frac{1}{2}\sum\limits_{i=1}^{I-1}\br{\alpha_n^{i+1} - \alpha_n^{i}
		- \frac{\delta_n}{\beta^{i}}}\norm{\u_n^* - \v_n^i}^2}
\end{split}
\end{equation}
\hrulefill
\end{figure*}

Based on \eqref{apdx6:01}, we introduce the following constraints in order to guarantee the convergence of LFRPG
\begin{subequations}\label{apdx6:02}
\begin{align}
\frac{\delta_0}{\beta^{i}} - \frac{\epsilon}{\beta^{i+1}} &> 0 \label{apdx6:02a}\\
\frac{\delta_0}{\beta^{i}} - \frac{\epsilon}{\beta^{i+1}} &\ge \alpha_0^{i+1} - \alpha_0^{i} \label{apdx6:02b}\\
\frac{\delta_n}{\beta^{i}} &\ge \alpha_n^{i+1} - \alpha_n^{i}, n = 1, \ldots, N \label{apdx6:02c}\\
\alpha_0^i &= \frac{3}{2}\br{\frac{c_0}{\br{\beta^i}^2} + L_0} \label{apdx6:02d}\\
\alpha_n^i &= \frac{c_n}{\br{\beta^i}^2} + L_n, n = 1, \ldots, N. \label{apdx6:02e}
\end{align}
\end{subequations}

Equation \eqref{apdx6:02a} implies that $\epsilon < \frac{3}{4}\delta_0$, based on which we select $\epsilon = \frac{1}{2}\delta_0$. Using \eqref{apdx6:02b} and \eqref{apdx6:02d}, we obtain $c_0 \le \frac{4}{21}\delta_0$; and based on \eqref{apdx6:02c} and \eqref{apdx6:02e}, we find $c_n \le \frac{6}{7}\delta_n$. Thus, we set the stepsize $\alpha_n^i$ as
\begin{equation}\label{apdx6:03}
\alpha_{n}^{i} =
\left\{\begin{array}{l}
\frac{\delta_0}{14}\br{i+2}^2 + \frac{3}{2}L_0, n = 0\\
\frac{3\delta_n}{14}\br{i+2}^2 + L_n, n = 1, \ldots, N.
\end{array}\right.
\end{equation}

We now can establish convergence of $\bar\w^I = T^{-1}\sum\nolimits_{k=1}^T\w_k^I$ as 
\begin{equation}\label{apdx6:04}
\begin{split}
& F\br{\bar\w^I} - F\br{\u^*} \\
\le & \frac{1}{T}\sumkt F\br{\w_k^I} - F\br{\u^*} \\
\le &  \frac{2 \eta_{16} }{T\br{I+2}^2}
 + \frac{ \eta_{17} }{\br{I+2}^2}
 + \frac{ I \eta_{18} }{\br{I+2}^2}
+ \OO{\frac{\lambda^2B^2 G}{\delta_0}}
\end{split}
\end{equation}
where $\eta_{16}$, $\eta_{17}$ and $\eta_{18}$ are defined respectively as
\begin{align}
\eta_{16} :=  & {\summ\alpha_n^1\norm{u_n^* - v_n^0}^2 } \label{apdx6:05a}\\
\eta_{17} :=  & {\br{\frac{3}{2}\delta_0 + 2\alpha_0^1}\norm{u_0^* - v_0^0}^2
 + \frac{4}{T}\sumkt F(\w_k^{0}) - 4 F\br{\u^*}} \label{apdx6:05b}\\
\eta_{18} := & {\summ\frac{7\sigma_n^2}{3\delta_n} + \frac{11\sigma_0^2 + 28\lambda^2Q^2 G}{\delta_0}}. \label{apdx6:05c}
\end{align}

\bibliographystyle{IEEEtran}
\bibliography{mach_learn_dyj}

\begin{thebibliography}{10}
\providecommand{\url}[1]{#1}
\csname url@samestyle\endcsname
\providecommand{\newblock}{\relax}
\providecommand{\bibinfo}[2]{#2}
\providecommand{\BIBentrySTDinterwordspacing}{\spaceskip=0pt\relax}
\providecommand{\BIBentryALTinterwordstretchfactor}{4}
\providecommand{\BIBentryALTinterwordspacing}{\spaceskip=\fontdimen2\font plus
\BIBentryALTinterwordstretchfactor\fontdimen3\font minus
  \fontdimen4\font\relax}
\providecommand{\BIBforeignlanguage}[2]{{%
\expandafter\ifx\csname l@#1\endcsname\relax
\typeout{** WARNING: IEEEtran.bst: No hyphenation pattern has been}%
\typeout{** loaded for the language `#1'. Using the pattern for}%
\typeout{** the default language instead.}%
\else
\language=\csname l@#1\endcsname
\fi
#2}}
\providecommand{\BIBdecl}{\relax}
\BIBdecl

\bibitem{Li2019a}
B.~Li, M.~Ma, and G.~B. Giannakis, ``On the convergence of {SARAH }and
  beyond,'' in \emph{Proc. International Conference on Artificial Intelligence
  and Statistics (AISTAT)}, vol. 108, Aug. 2020, pp. 223--233.

\bibitem{Li2019b}
B.~Li, L.~Wang, and G.~B. Giannakis, ``Almost tune-free variance reduction,''
  \emph{Proc. International Conference on Machine Learning (ICML)}, 2020.

\bibitem{Konecny2015}
J.~Kone$\check{\mbox{c}}$n{\'y}, B.~McMahan, and D.~Ramage, ``Federated
  optimization: Distributed optimization beyond the datacenter,'' \emph{arXiv
  preprint arXiv:1511.03575}, Mar. 2015.

\bibitem{Dong2019}
Y.~Dong, J.~Cheng, M.~J. Hossain, and V.~C.~M. Leung, ``Secure distributed
  on-device learning networks with {B}yzantine adversaries,'' \emph{IEEE
  Netw.}, vol.~33, no.~6, pp. 180--187, Nov.--Dec. 2019.

\bibitem{chen2019joint}
M.~Chen, Z.~Yang, W.~Saad, C.~Yin, H.~V. Poor, and S.~Cui, ``A joint learning
  and communications framework for federated learning over wireless networks,''
  \emph{arXiv preprint arXiv:1909.07972}, Sept. 2019.

\bibitem{9134426}
Y.~{Shi}, K.~{Yang}, T.~{Jiang}, J.~{Zhang}, and K.~B. {Letaief},
  ``Communication-efficient edge {AI}: Algorithms and systems,'' \emph{IEEE
  Commun. Surveys Tut.}, to be published, 2020.

\bibitem{chen2017}
Y.~Chen, L.~Su, and J.~Xu, ``Distributed statistical machine learning in
  adversarial settings: Byzantine gradient descent,'' \emph{Proc. ACM Meas.
  Anal. Comput. Syst.}, vol.~1, no.~2, pp. 44:1--44:25, Dec. 2017.

\bibitem{pmlr-v80-yin18a}
D.~Yin, Y.~Chen, R.~Kannan, and P.~Bartlett, ``{B}yzantine-robust distributed
  learning: Towards optimal statistical rates,'' in \emph{Proc. International
  Conference on Machine Learning (ICML)}, Stockholmsm\"{a}ssan, Stockholm,
  Sweden, July 2018, pp. 5650--5659.

\bibitem{pmlr-v97-yin19a}
------, ``Defending against saddle point attack in {B}yzantine-robust
  distributed learning,'' in \emph{Proc. International Conference on Machine
  Learning (ICML)}, vol.~97, Long Beach, California, USA, June 2019, pp.
  7074--7084.

\bibitem{su2019}
L.~Su and J.~Xu, ``Securing distributed gradient descent in high dimensional
  statistical learning,'' in \emph{Proc. ACM Meas. Anal. Comput. Syst.},
  vol.~3, no.~1, Mar. 2019, pp. 12:1--12:41.

\bibitem{NIPS2017_6617}
P.~Blanchard, E.~M. El~Mhamdi, R.~Guerraoui, and J.~Stainer, ``Machine learning
  with adversaries: Byzantine tolerant gradient descent,'' in \emph{Proc.
  Advances in Neural Information Processing Systems (NIPS)}, Long Beach, USA,
  Dec. 2017, pp. 119--129.

\bibitem{pmlr-v80-mhamdi18a}
E.~M. El~Mhamdi, R.~Guerraoui, and S.~Rouault, ``The hidden vulnerability of
  distributed learning in {B}yzantium,'' in \emph{Proc. International
  Conference on Machine Learning (ICML)}, Stockholmsm\"{a}ssan, Stockholm,
  Sweden, July 2018, pp. 3521--3530.

\bibitem{Alistarh2018}
D.~Alistarh, Z.~Allen-Zhu, and J.~Li, ``Byzantine stochastic gradient
  descent,'' in \emph{Proc. Advances in Neural Information Processing Systems
  (NIPS)}, Montreal, CA, Dec. 2018, pp. 4614--4624.

\bibitem{pmlr-v97-xie19b}
C.~Xie, S.~Koyejo, and I.~Gupta, ``Zeno: Distributed stochastic gradient
  descent with suspicion-based fault-tolerance,'' in \emph{Proc. International
  Conference on Machine Learning (ICML)}, Long Beach, California, USA, June
  2019, pp. 6893--6901.

\bibitem{Chen2018}
L.~Chen, H.~Wang, Z.~Charles, and D.~Papailiopoulos, ``{DRACO}:
  {B}yzantine-resilient distributed training via redundant gradients,'' in
  \emph{Proc. International Conference on Machine Learning (ICML)},
  Stockholmsm\"{a}ssan, Stockholm, Sweden, July 2018, pp. 903--912.

\bibitem{xie_slsgd}
C.~Xie, O.~Koyejo, and I.~Gupta, ``Slsgd: Secure and efficient distributed
  on-device machine learning,'' in \emph{Machine Learning and Knowledge
  Discovery in Databases}, 2020, pp. 213--228.

\bibitem{Li2019}
L.~Li, W.~Xu, T.~Chen, G.~B. Giannakis, and Q.~Ling, ``{RSA}:
  {Byzantine}-robust stochastic aggregation methods for distributed learning
  from heterogeneous datasets,'' in \emph{Proc. AAAI Conference on Artificial
  Intelligence}, vol.~33, no.~01, Jan. 2019, pp. 1544--1551.

\bibitem{Ghosh2019}
A.~Ghosh, J.~Hong, D.~Yin, and K.~Ramchandran, ``Robust federated learning in a
  heterogeneous environment,'' \emph{arXiv preprint arXiv:1906.06629}, 2019.

\bibitem{NIPS2014_5597}
M.~Li, D.~G. Andersen, A.~J. Smola, and K.~Yu, ``Communication efficient
  distributed machine learning with the parameter server,'' in \emph{Proc.
  Advances in Neural Information Processing Systems (NIPS)}, Palais des
  Congr$\grave{\mbox{e}}$s de Montr$\acute{\mbox{e}}$al,
  Montr$\acute{\mbox{e}}$al, Dec. 2014, pp. 19--27.

\bibitem{michaeljason2019}
M.~I. Jordan, J.~D. Lee, and Y.~Yang, ``Communication-efficient distributed
  statistical inference,'' \emph{Journal of the American Statistical
  Association}, vol. 114, no. 526, pp. 668--681, 2019.

\bibitem{8889996}
F.~{Sattler}, S.~{Wiedemann}, K.~{Müller}, and W.~{Samek}, ``Robust and
  communication-efficient federated learning from non-i.i.d. data,'' \emph{IEEE
  Trans. Neural Netw. Learn. Syst.}, to be published, 2019.

\bibitem{Chentobepublished}
T.~Chen, G.~B. Giannakis, T.~Sun, and W.~Yin, ``{LAG}: Lazily aggregated
  gradient for communication-efficient distributed learning,'' in \emph{Proc.
  Advances in Neural Information Processing Systems (NIPS)}, Montreal, CA, Dec.
  2018, pp. 5050--5060.

\bibitem{Sun2019}
J.~Sun, T.~Chen, G.~B. Giannakis, and Z.~Yang, ``Communication-efficient
  distributed learning via lazily aggregated quantized gradients,'' in
  \emph{Proc. Advances in Neural Information Processing Systems (NIPS)}, to be
  published, Sept. 2019.

\bibitem{Stich2019}
S.~U. Stich, ``Local {SGD} converges fast and communicates little,'' in
  \emph{Proc. International Conference on Learning Representations (ICLR)},
  Addis Ababa, Ethiopia, Apr. 2019.

\bibitem{yu2018parallel}
H.~Yu, S.~Yang, and S.~Zhu, ``Parallel restarted {SGD} with faster convergence
  and less communication: Demystifying why model averaging works for deep
  learning,'' in \emph{Proc. AAAI Conference on Artificial Intelligence},
  vol.~33, no.~01, Jan. 2019, pp. 5693--5700.

\bibitem{pmlr-v97-yu19d}
H.~Yu, R.~Jin, and S.~Yang, ``On the linear speedup analysis of communication
  efficient momentum {SGD} for distributed non-convex optimization,'' in
  \emph{Proc. International Conference on Machine Learning (ICML)}, vol.~97,
  Long Beach, California, USA, June 2019, pp. 7184--7193.

\bibitem{khaled2019first}
A.~Khaled, K.~Mishchenko, and P.~Richt{\'a}rik, ``First analysis of local {GD
  }on heterogeneous data,'' \emph{arXiv preprint arXiv:1909.04715}, 2019.

\bibitem{Nesterov:2014:ILC:2670022}
Y.~Nesterov, \emph{Introductory Lectures on Convex Optimization: A Basic
  Course}, 1st~ed.\hskip 1em plus 0.5em minus 0.4em\relax Springer Publishing
  Company, Incorporated, 2014.

\bibitem{Hu2009}
C.~Hu, W.~Pan, and J.~T. Kwok, ``Accelerated gradient methods for stochastic
  optimization and online learning,'' in \emph{Proc. Advances in Neural
  Information Processing Systems (NIPS)}, Vancouver, Canada, Dec. 2009, pp.
  781--789.

\bibitem{KoppelJune2017}
A.~Koppel, B.~M. Sadler, and A.~Ribeiro, ``Proximity without consensus in
  online multiagent optimization,'' \emph{IEEE Trans. Signal Process.},
  vol.~65, no.~12, pp. 3062--3077, June 2017.

\bibitem{latorre2020lipschitz}
F.~Latorre, P.~Rolland, and V.~Cevher, ``Lipschitz constant estimation of
  neural networks via sparse polynomial optimization,'' \emph{Proc.
  International Conference on Learning Representations (ICLR)}, to be
  published, 2020.

\bibitem{Nemirovski2009}
A.~Nemirovski, A.~Juditsky, G.~Lan, and A.~Shapiro, ``Robust stochastic
  approximation approach to stochastic programming,'' \emph{SIAM J. Opt.},
  vol.~19, no.~4, pp. 1574--1609, 2009.

\bibitem{291440}
\BIBentryALTinterwordspacing
J.~J. {Hull}, ``A database for handwritten text recognition research,''
  \emph{IEEE Trans. Pattern Anal. Mach. Intell.}, vol.~16, no.~5, pp. 550--554,
  May 1994. [Online]. Available: \url{https://cs.nyu.edu/~roweis/data.html}
\BIBentrySTDinterwordspacing

\bibitem{726791}
\BIBentryALTinterwordspacing
Y.~{Lecun}, L.~{Bottou}, Y.~{Bengio}, and P.~{Haffner}, ``Gradient-based
  learning applied to document recognition,'' \emph{Proc. IEEE}, vol.~86,
  no.~11, pp. 2278--2324, Nov. 1998. [Online]. Available:
  \url{http://yann.lecun.com/exdb/mnist/}
\BIBentrySTDinterwordspacing

\bibitem{xiao2017fashion}
\BIBentryALTinterwordspacing
H.~Xiao, K.~Rasul, and R.~Vollgraf, ``Fashion-{MNIST}: a novel image dataset
  for benchmarking machine learning algorithms,'' \emph{arXiv preprint
  arXiv:1708.07747}, Sept. 2017. [Online]. Available:
  \url{https://www.kaggle.com/koushikk/fmnist}
\BIBentrySTDinterwordspacing

\bibitem{weiszfeld2009point}
E.~Weiszfeld and F.~Plastria, ``On the point for which the sum of the distances
  to n given points is minimum,'' \emph{Annals of Operations Research}, vol.
  167, no.~1, pp. 7--41, 2009.

\bibitem{Beckenbach2012}
E.~F. Beckenbach and R.~Bellman, \emph{Inequalities}.\hskip 1em plus 0.5em
  minus 0.4em\relax Springer Science \& Business Media, 2012, vol.~30.

\end{thebibliography}

\clearpage
\onecolumn
\begin{center}
{\Large\bf Supplementary Materials for ``Communication-Efficient Robust Federated Learning''}
\end{center}

\begin{center}
{\bf Yanjie Dong, Gerogios B. Giannakis, Tianyi Chen, Julian Cheng, Md. Jahangir Hossain, and Victor C. M. Leung}
\end{center}

\section{Proof of Lemma \ref{le:06}}\label{apdx:05}
Since $u_0^i$ and $w_0^{i}$ are updated at the start of frame $i$, we set $u_0^i = u_{0,k}^i$ and $w_0^{i} = w_{0,k}^{i}$ for $k = 1, \ldots, T$, and also define $h_{0,k}^i := \alpha_0^i({u_0^i - w_0^i})$ and $h_{n,k}^i := \alpha_{n,k}^i({u_{n,k}^i - w_{n,k}^i})$.
With $v_{n,k}^i$ denoting the minimizer of
\begin{equation}\label{apdx5:01}
\phi_{n,k}^i :=  \inp{h_{n,k}^i, u_n - u_{n,k}^i}
+ \frac{\delta_n}{2}\norm{u_n - u_{n,k}^i}^2
 + \frac{\alpha_n^{i}\beta^{i}}{2}\norm{u_n - v_{n, k-1}^{i}}^2,~~~n=1,\ldots,N
\end{equation}
it holds that 
\begin{equation}\label{apdx5:02}
\begin{split}
& \inp{h_{n,k}^i, u_n - u_{n,k}^i} + \frac{\delta_n}{2}\norm{u_n - u_{n,k}^i}^2 \\
\ge& \inp{h_{n,k}^i, v_{n,k}^i - u_{n,k}^i}
 + \frac{\delta_n + \alpha_n^{i}\beta^{i}}{2}\norm{u_n - v_{n,k}^i}^2 
 + \frac{\alpha_n^{i}\beta^{i}}{2}\norm{v_{n,k}^i - v_{n, k-1}^i}^2
- \frac{\alpha_n^{i}\beta^{i}}{2}\norm{u_n - v_{n,k-1}^i}^2.
\end{split}
\end{equation}
Upon substituting \eqref{apdx5:02} into \eqref{eqa:lbrpg:05}, we have
\begin{equation}\label{apdx5:03}
\begin{split}
 F\br{\w_k^i} - F\br{\u}
\le & \sumam\inp{\Delta_{n,k}^i, u_n - w_{n,k}^{i}}
+ \frac{\norm{\sumq g_{n,k}^i}}{\alpha_0^{i}}\norm{h_{0,k}^{i}}
- \sumam\frac{2\alpha_n^{i} - L_n}{2\br{\alpha_n^{i}}^2}\norm{h_{n,k}^i}^2 \\
& - \inp{h_{0,k}^{i} + \sumq g_{n,k}^i, u_0 - u_{0, k}^{i}}
- \frac{\delta_0}{2}\norm{u_0 - u_{0,k}^i}^2
 + \summ\inp{h_{n,k}^i, u_{n,k}^i - v_{n,k}^i}
 + \summ \frac{ \eta_{11, n, k}^i }{\beta^{i}}.
\end{split}
\end{equation}
where 
\begin{equation}\label{apdx5:04}
\eta_{11, n, k}^i :=  \frac{\alpha_n^{i}\br{\beta^{i}}^2}{2}\norm{u_n - v_{n,k-1}^i}^2
 - \frac{\delta_n\beta^{i} + \alpha_n^{i}\br{\beta^{i}}^2}{2}\norm{u_n - v_{n,k}^i}^2
 - \frac{\alpha_n^{i}\br{\beta^{i}}^2}{2}\norm{v_{n,k}^i - v_{n,k-1}^i}^2.
\end{equation}
Setting $\u = \w_k^{i-1}$ into \eqref{eqa:lbrpg:05} and dropping the non-positive term $-\sum\nolimits_{n=0}^N\frac{\delta_n}{2}\norm{w_{n,k}^i - u_{n,k}^i}^2$, we obtain
\begin{equation}\label{apdx5:05}
\begin{split}
 F\br{w_k^i} - F\br{w_k^{i-1}}
\le & \sumam\inp{\Delta_{n,k}^i, w_{n,k}^{i-1} - w_{n,k}^{i}}
+ \frac{\norm{\sumq g_{n,k}^i}}{\alpha_0^{i}}\norm{h_{0,k}^{i}} \\
& - \sumam\frac{2\alpha_n^{i} - L_n}{2\br{\alpha_n^{i}}^2}\norm{h_{n,k}^i}^2
 - \inp{h_{0,k}^{i} + \sumq g_{n,k}^i, w_{0,k}^{i-1} - u_{0, k}^{i}}
 - \summ\inp{h_{n,k}^i, w_{n,k}^{i-1} - u_{n,k}^i}.
\end{split}
\end{equation}
Consider for future use the convex combination of \eqref{apdx5:03} and \eqref{apdx5:05} as
\begin{equation}\label{apdx5:06}
\beta^{i}\br{F\br{\w_k^i} - F\br{\u} } + \br{1-\beta^{i}}\br{ F\br{\w_k^i} - F\br{\w_k^{i-1}}}
\end{equation}
and upper bound \eqref{apdx5:06} as
\begin{equation}\label{apdx5:07}
\begin{split}
& F\br{\w_k^i} - F\br{\u} \\
\le & \br{1-\beta^{i}}\br{ F\br{\w_k^{i-1}} - F\br{\u} }
 + \frac{\norm{\sumq g_{n,k}^i}}{\alpha_0^{i}}\norm{h_{0,k}^{i}}
  - \sumam\frac{2\alpha_n^{i} - L_n}{2\br{\alpha_n^{i}}^2}\norm{h_{n,k}^i}^2  + \summ \br{ \eta_{11, n, k}^i + \eta_{12, n, k}^i + \eta_{13, n, k}^i } \\
&+ \eta_{12, 0, k}^i + \beta^i\inp{h_{0,k}^i+\sumq g_{n,k}^i, u_{0,k}^i - u_0} - \frac{\delta_0\beta^i}{2}\norm{u_0 - u_{0,k}^i}^2
 + \br{1-\beta^i}\inp{h_{0,k}^i+\sumq g_{n,k}^i, u_{0,k}^i - w_{0,k}^{i-1}}
\end{split}
\end{equation}
where $\eta_{12, n, k}^i$ and $\eta_{13, n, k}^i$ are respectively defined as
\begin{align}
%\eta_{11, n, k}^i :=& \inp{\Delta_{n,k}^i, \beta^i u_n + \br{1-\beta^i} w_{n,k}^{i-1} -  w_{n,k}^i} \label{apdx5:08}\\
\eta_{12, n, k}^i :=& \inp{\Delta_{0,k}^i, \beta^{i} u_0 + \br{1-\beta^{i}} w_{0,k}^{i-1} - w_{0,k}^{i} }, n = 0, 1, \ldots, N \mbox{ and } \label{apdx5:08}\\
\eta_{13, n, k}^i :=& \inp{ h_{n,k}^i,  u_{n,k}^i - \beta^{i} v_{n,k}^i - \br{1-\beta^{i}} w_{n,k}^{i-1}}. \label{apdx5:09}
\end{align}

Based on \eqref{eqa:lbrpg:02a}, we simplify \eqref{apdx5:08} for $n = 1,\ldots, N$ and leverage H\"{o}lder's inequality \cite{Beckenbach2012}, to write
\begin{equation}\label{apdx5:11}
\eta_{12, n, k}^i
= \beta^{i}\inp{\Delta_{n,k}^i, u_n - v_{n,k-1}^i}
+ \inp{\Delta_{n,k}^i, u_{n,k}^i - w_{n,k}^i} 
\le \beta^{i}\inp{\Delta_{n,k}^i, u_n - v_{n,k-1}^i}  + \frac{\sqrt{\norm{\Delta_{n,k}^i}^2}}{\alpha_n^i}\norm{h_{n,k}^i}.
\end{equation}
Using arguments similar to those in \eqref{apdx3:13}--\eqref{apdx3:15}, we take expectation over both sides of \eqref{apdx5:11} to arrive at 
\begin{equation}\label{apdx5:12}
\E_{\x_{n,k}^{1:i}}\sq{ \eta_{12, n, k}^i } \le \frac{\sigma_n}{\alpha_{n}^{i}}\norm{h_{n,k}^i}.
\end{equation}

Substituting \eqref{apdx5:12} into \eqref{apdx5:07}, we have
\begin{equation}\label{apdx5:13}
\begin{split}
 F\br{\w_k^i} - F\br{\u}
\le & \br{1-\beta^{i}}\br{ F\br{\w_k^{i-1}} - F\br{\u} }
 + \frac{\norm{\sumq g_{n,k}^i}}{\alpha_0^{i}}\norm{h_{0,k}^{i}}
  - \sumam\frac{2\alpha_n^{i} - L_n}{2\br{\alpha_n^{i}}^2}\norm{h_{n,k}^i}^2  \\
& + \summ\frac{\sigma_n}{\alpha_n^i}\norm{h_{n,k}^i}
+ \summ \br{ \eta_{11, n, k}^i + \eta_{13, n, k}^i } 
+ \eta_{12, 0, k}^i  \\
& + \beta^i\inp{h_{0,k}^i+\sumq g_{n,k}^i, u_{0,k}^i - u_0} - \frac{\delta_0\beta^i}{2}\norm{u_0 - u_{0,k}^i}^2
 + \br{1-\beta^i}\inp{h_{0,k}^i+\sumq g_{n,k}^i, u_{0,k}^i - w_{0,k}^{i-1}}.
\end{split}
\end{equation}
Since $u_{0,k}^i$ and $w_{0,k}^i$ remain constant over frame $i$, it follows that $h_{0,k}^i = h_0^i$. Summing \eqref{apdx5:13} over $k = 1, \ldots, T$ and dividing by $T$, we have
\begin{equation}\label{apdx5:14}
\begin{split}
&\frac{1}{T}\sumkt F\br{\w_k^i} - F\br{\u}  \\
\le & \br{1-\beta^{i}}\br{ \frac{1}{T}\sumkt F\br{\w_k^{i-1}} - F\br{\u} }
 + \frac{\frac{1}{T}\sumkt\norm{\sumq g_{n,k}^i}}{\alpha_0^{i}}\norm{h_{0}^{i}}
 + \frac{1}{T}\sumkt\summ\frac{\sigma_n}{\alpha_n^i}\norm{h_{n,k}^i}
 \\
&- \frac{2\alpha_0^{i} - L_0}{2\br{\alpha_0^{i}}^2}\norm{h_{0}^i}^2 
- \frac{1}{T}\sumkt\summ\frac{2\alpha_n^{i} - L_n}{2\br{\alpha_n^{i}}^2}\norm{h_{n,k}^i}^2
+ \frac{1}{T}\sumkt\summ \br{ \eta_{11, n, k}^i + \eta_{13, n, k}^i }
+ \frac{1}{T}\sumkt \eta_{12, 0, k}^i \\
& + \beta^i\inp{h_{0}^i+\frac{1}{T}\sumkt\sumq g_{n,k}^i,  u_{0}^i -  u_0} - \frac{\delta_0\beta^i}{2}\norm{u_0 - u_{0}^i}^2
 + \br{1-\beta^i}\inp{h_{0}^i+\frac{1}{T}\sumkt\sumq g_{n,k}^i, u_{0}^i - w_{0}^{i-1}}.
\end{split}
\end{equation}
Recall also that $v_0^i$ is the minimizer of $\phi_0^{i}\br{u_0}$ according to \eqref{eqa:lbrpg:01c}, where $\phi_0^{i}\br{u_0}$ is defined as
\begin{equation}\label{apdx5:15}
\phi_0^{i}\br{ u_0} := \inp{h_0^{i} + \frac{1}{T}\sumkt\sumq g_{n,k}^i, u_0 - u_0^{i}}
 + \frac{\delta_0}{2}\norm{u_0 - u_0^{i}}^2
+ \frac{\alpha_0^{i}\beta^{i}}{2}\norm{ u_0 -  v_0^{i-1}}^2.
\end{equation}
Since $\phi_0^{i}\br{\u_0}$ in \eqref{apdx5:15} is strongly convex with modulus $\delta_0 + \alpha_0^{i}\beta^{i}$, we deduce that 
\begin{equation}\label{apdx5:16}
\begin{split}
\inp{ h_0^{i} + \frac{1}{T}\sumkt\sumq g_{n,k}^i, u_0 - u_0^{i}}
+ \frac{\delta_0}{2}\norm{ u_0 -  u_0^{i}}^2
\ge & \inp{ h_0^{i} + \frac{1}{T}\sumkt\sumq g_{n,k}^i, v_0^{i} - u_0^{i}}
+ \frac{\delta_0 + \alpha_0^{i}\beta^{i}}{2}\norm{u_0 - v_0^{i}}^2 \\
&+ \frac{\alpha_0^{i}\beta^{i}}{2}\norm{ v_0^{i} -  v_0^{i-1}}^2
 - \frac{\alpha_0^{i}\beta^{i}}{2}\norm{ u_0 -  v_0^{i-1}}^2.
\end{split}
\end{equation}
Substituting \eqref{apdx5:16} into \eqref{apdx5:14}, it follows after straightforward  manipulations using \eqref{eqa:lbrpg:01a}, that 
\begin{equation}\label{apdx5:17}
\begin{split}
& \frac{1}{T}\sumkt F\br{\w_k^i} - F\br{\u}\\
\le & \br{1-\beta^{i}}\br{ \frac{1}{T}\sumkt F\br{\w_k^{i-1}} - F\br{\u} }
 + \frac{\frac{1}{T}\sumkt\norm{\sumq g_{n,k}^i}}{\alpha_0^{i}}\norm{ h_{0}^{i}}
 + \frac{1}{T}\sumkt\summ\frac{\sigma_n}{\alpha_n^i}\norm{ h_{n,k}^i}
- \frac{2\alpha_0^{i} - L_0}{2\br{\alpha_0^{i}}^2}\norm{ h_{0}^i}^2 \\
&- \frac{1}{T}\sumkt\summ\frac{2\alpha_n^{i} - L_n}{2\br{\alpha_n^{i}}^2}\norm{ h_{n,k}^i}^2
+ \frac{1}{T}\sumkt\summ \br{ \eta_{11, n, k}^i  + \eta_{13, n, k}^i }
+ \frac{1}{T}\sumkt \eta_{12, 0, k}^i
+ \beta^i\inp{h_0^{i} + \frac{1}{T}\sumkt\sumq g_{n,k}^i,  v_0^{i-1} -  v_0^{i}} \\
& + \frac{\alpha_0^{i}\br{\beta^{i}}^2}{2}\norm{ u_0 - v_0^{i-1}}^2
- \frac{\delta_0\beta^{i} + \alpha_0^{i}\br{\beta^{i}}^2}{2}\norm{ u_0 - v_0^{i}}^2
- \frac{\alpha_0^{i}\br{\beta^{i}}^2}{2}\norm{ v_0^{i} -  v_0^{i-1}}^2 \\
\le & \br{1-\beta^{i}}\br{ \frac{1}{T}\sumkt F\br{\w_k^{i-1}} - F\br{\u} }
 + \frac{\frac{1}{T}\sumkt\norm{\sumq g_{n,k}^i}}{\alpha_0^{i}}\norm{ h_{0}^{i}}
 + \frac{1}{T}\sumkt\summ\frac{\sigma_n}{\alpha_n^i}\norm{ h_{n,k}^i}
- \frac{2\alpha_0^{i} - L_0}{2\br{\alpha_0^{i}}^2}\norm{ h_{0}^i}^2 \\
&- \frac{1}{T}\sumkt\summ\frac{2\alpha_n^{i} - L_n}{2\br{\alpha_n^{i}}^2}\norm{ h_{n,k}^i}^2
+ \frac{1}{T}\sumkt\summ \br{ \eta_{11, n, k}^i + \eta_{13, n, k}^i }
+ \frac{1}{T}\sumkt \eta_{12, 0, k}^i
+ \frac{1}{2\alpha_0^i}\norm{ h_0^{i} + \frac{1}{T}\sumkt\sumq g_{n,k}^i}^2\\
& + \frac{\alpha_0^{i}\br{\beta^{i}}^2}{2}\norm{ u_0 -  v_0^{i-1}}^2
- \frac{\delta_0\beta^{i} + \alpha_0^{i}\br{\beta^{i}}^2}{2}\norm{ u_0 -  v_0^{i}}^2
\end{split}
\end{equation}
where the second inequality follows from Young's inequality \cite{Beckenbach2012}.

The convex term $\|{h_0^{i} + \frac{1}{T}\sum\nolimits_{k=1}^T\sum\nolimits_{n=1}^Q g_{n,k}^i}\|^2$ is upper-bounded by
\begin{equation}\label{apdx5:18}
\norm{h_0^{i} + \frac{1}{T}\sumkt\sumq g_{n,k}^i}^2 \le
\frac{4}{3}\norm{h_0^i}^2
+ 4\norm{\frac{1}{T}\sumkt\sumq g_{n,k}^i}^2 
\le \frac{4}{3}\norm{h_0^i}^2
+ 4\lambda^2Q^2 G
\end{equation}
where the second inequality holds because $\|{\frac{1}{T}\sum\nolimits_{k=1}^T\sum\nolimits_{n=1}^Q g_{n,k}^i}\|^2 \le \lambda^2Q^2 G$.

Based on \eqref{eqa:lbrpg:02a}, we simplify $\eta_{13, n, k}^i$, and use Young's inequality \cite{Beckenbach2012}, to write 
\begin{equation}\label{apdx5:19}
\eta_{13, n, k}^i = \beta^{i} \inp{ h_{n,k}^i,  v_{n,k-1}^i -  v_{n,k}^i } 
\le \frac{\norm{ h_{n,k}^i}^2}{2\alpha_{n}^{i}}
+ \frac{\alpha_{n}^{i}\br{\beta^{i}}^2}{2}\norm{ v_{n,k-1}^i -  v_{n,k}^i}^2\:.
\end{equation}
Next, we simplify $\eta_{12, 0, k}^i$ using \eqref{eqa:lbrpg:02a}, as
\begin{equation}\label{apdx5:20}
\begin{split}
\eta_{12, 0, k}^i =& \inp{\Delta_{0,k}^i, \beta^{i} u_0  - \beta^{i} v_{0,k}^{i-1} +  u_{0}^{i} -  w_{0}^{i} } \\
\le & \beta^{i}\inp{\Delta_{0,k}^i,  u_0 -  v_{0,k}^{i-1}}
+ \frac{\norm{\Delta_{0,k}^i}}{\alpha_0^{i}}\norm{ h_{0}^{i}} \\
=& \beta^{i}\inp{\Delta_{0,k}^i,  u_0 -  v_{0}^{i-1}}
+ \frac{\norm{\Delta_{0,k}^i}}{\alpha_0^{i}}\norm{ h_{0}^{i}} \\
\le & \frac{\beta^i}{2\epsilon}\norm{\Delta_{0,k}^i}^2
+ \frac{\epsilon\beta^i}{2}\norm{ u_0 -  v_{0}^{i-1}}^2
+ \frac{\norm{\Delta_{0,k}^i}}{\alpha_0^{i}}\norm{ h_{0}^{i}} \\
\le& \frac{\beta^i}{2\epsilon}\lambda^2B^2 G
+ \frac{\epsilon\beta^i}{2}\norm{ u_0 -  v_{0}^{i-1}}^2
+ \frac{\norm{\Delta_{0,k}^i}}{\alpha_0^{i}}\norm{ h_{0}^{i}}
\end{split}
\end{equation}
where the first inequality is due to H\"{o}lder's inequality \cite{Beckenbach2012}, the second follows from Young's inequality \cite{Beckenbach2012}, and the third relies on the fact that  $\|{\Delta_{0,k}^i}\|^2 \le \lambda^2B^2 G$.

Substituting \eqref{apdx5:18}--\eqref{apdx5:20} into \eqref{apdx5:17}, we obtain
\begin{equation}\label{apdx5:21}
\begin{split}
& \frac{1}{T}\sumkt F\br{\w_k^i} - F\br{\u}\\
\le & \br{1-\beta^{i}}\br{ \frac{1}{T}\sumkt F\br{\w_k^{i-1}} - F\br{\u} }
 + \frac{\frac{1}{T}\sumkt\br{\norm{\sumq g_{n,k}^i} + \norm{\Delta_{0,k}^i}}}{\alpha_0^{i}}\norm{h_{0}^{i}}
- \frac{2\alpha_0^{i} - 3 L_0}{6 \br{\alpha_0^{i}}^2}\norm{h_{0}^i}^2 \\
&+ \frac{\beta^i}{2\epsilon}\lambda^2B^2 G
+ \frac{2\lambda^2Q^2G}{\alpha_0^i}+ \frac{1}{T}\sumkt\summ\frac{\sigma_n}{\alpha_n^i}\norm{ h_{n,k}^i}
- \frac{1}{T}\sumkt\summ\frac{\alpha_n^{i} - L_n}{2\br{\alpha_n^{i}}^2}\norm{ h_{n,k}^i}^2 + \frac{\epsilon\beta^i + \alpha_0^{i}\br{\beta^{i}}^2}{2}\norm{ u_0 - v_0^{i-1}}^2 \\
&
- \frac{\delta_0\beta^{i} + \alpha_0^{i}\br{\beta^{i}}^2}{2}\norm{ u_0 - v_0^{i}}^2  + \frac{1}{T}\summ\sumkt\br{\frac{\alpha_n^i\br{\beta^i}^2}{2}\norm{ u_n - v_{n,k-1}^i}^2 -\frac{\delta_n\beta^i + \alpha_n^i\br{\beta^i}^2}{2}\norm{ u_n - v_{n,k}^i}^2 } \\
\le & \br{1-\beta^{i}}\br{ \frac{1}{T}\sumkt F\br{\w_k^{i-1}} - F\br{\u} }
 + \frac{\frac{1}{T}\sumkt\br{\norm{\sumq g_{n,k}^i} + \norm{\Delta_{0,k}^i}}}{\alpha_0^{i}}\norm{ h_{0}^{i}}
- \frac{2\alpha_0^{i} - 3 L_0}{6 \br{\alpha_0^{i}}^2}\norm{ h_{0}^i}^2 \\
&+ \frac{1}{T}\sumkt\summ\frac{\sigma_n}{\alpha_n^i}\norm{ h_{n,k}^i}
- \frac{1}{T}\sumkt\summ\frac{\alpha_n^{i} - L_n}{2\br{\alpha_n^{i}}^2}\norm{ h_{n,k}^i}^2
+ \frac{\beta^i}{2\epsilon}\lambda^2B^2G
+ \frac{2\lambda^2Q^2 G}{\alpha_0^i}+ \frac{\epsilon\beta^i + \alpha_0^{i}\br{\beta^{i}}^2}{2}\norm{ u_0 - v_0^{i-1}}^2 \\
& - \frac{\delta_0\beta^{i} + \alpha_0^{i}\br{\beta^{i}}^2}{2}\norm{ u_0 - v_0^{i}}^2  + \frac{1}{T}\summ\br{\frac{\alpha_n^i\br{\beta^i}^2}{2}\norm{ u_n - v_{n,T}^{i-1}}^2 -\frac{\delta_n\beta^i + \alpha_n^i\br{\beta^i}^2}{2}\norm{ u_n - v_{n,T}^i}^2 } \\
\le & \br{1-\beta^{i}}\br{ \frac{1}{T}\sumkt F\br{\w_k^{i-1}} - F\br{\u} }
+ \frac{3\sigma_0^2}{2\br{2\alpha_0^{i} - 3L_0}}
+ \summ\frac{\sigma_n^2}{2\br{\alpha_n^i - L_n}} + \frac{\beta^i}{2\epsilon}\lambda^2B^2 G
+ \frac{2\lambda^2Q^2 G}{\alpha_0^i} \\
& + \frac{\epsilon\beta^i + \alpha_0^{i}\br{\beta^{i}}^2}{2}\norm{ u_0 - v_0^{i-1}}^2
- \frac{\delta_0\beta^{i} + \alpha_0^{i}\br{\beta^{i}}^2}{2}\norm{ u_0 - v_0^{i}}^2 \\
& + \frac{1}{T}\summ\br{\frac{\alpha_n^i\br{\beta^i}^2}{2}\norm{ u_n - v_{n,T}^{i-1}}^2 -\frac{\delta_n\beta^i + \alpha_n^i\br{\beta^i}^2}{2}\norm{ u_n - v_{n,T}^i}^2 } \\
\le & \br{1-\beta^{i}}\br{ \frac{1}{T}\sumkt F\br{\w_k^{i-1}} - F\br{\u} }
+ \frac{3\sigma_0^2 + 8\lambda^2Q^2 G}{2\br{2\alpha_0^{i} - 3 L_0}}
+ \summ\frac{\sigma_n^2}{2\br{\alpha_n^i - L_n}} + \frac{\beta^i}{2\epsilon}\lambda^2B^2 G \\
& + \frac{\epsilon\beta^i + \alpha_0^{i}\br{\beta^{i}}^2}{2}\norm{ u_0 - v_0^{i-1}}^2
- \frac{\delta_0\beta^{i} + \alpha_0^{i}\br{\beta^{i}}^2}{2}\norm{ u_0 - v_0^{i}}^2 \\
& + \frac{1}{T}\summ\br{\frac{\alpha_n^i\br{\beta^i}^2}{2}\norm{ u_n - v_{n,T}^{i-1}}^2 -\frac{\delta_n\beta^i + \alpha_n^i\br{\beta^i}^2}{2}\norm{ u_n - v_{n,T}^i}^2 }
\end{split}
\end{equation}
where the second inequality is obtained after setting $v_{n,0}^i = v_{n,T}^{i-1}$, and the third inequality follows because $-ax^2 + bx \le \frac{b^2}{4a}$ and  $\frac{1}{T^2}\br{\sum\nolimits_{k=1}^T\br{\|{\Delta_{0,k}^i}\| + \|{\sum\nolimits_{n=1}^Q g_{n,k}^i}}\| }^2 \le \sigma_0^2$.

Dividing both sides of \eqref{apdx5:21} by $\br{\beta^i}^2$ and setting $\u = \u^*$, we obtain
\begin{equation}\label{apdx5:22}
 \frac{1}{\br{\beta^i}^2}\br{\frac{1}{T}\sumkt F\br{\w_k^i} - F\br{\u^*}}
\le  \frac{1-\beta^{i}}{\br{\beta^i}^2}\br{ \frac{1}{T}\sumkt F\br{\w_k^{i-1}} - F\br{\u^*} }
+ \sumam \br{ \eta_{14, n}^i + \eta_{15, n}^i }
+ \frac{\lambda^2B^2 G  }{2\epsilon\beta^i}
\end{equation}
where 
\begin{equation}\label{apdx5:23}
\eta_{14, n}^i := \left\{\begin{array}{l}
\frac{3\sigma_0^2 + 8\lambda^2Q^2 G}{2\br{2\alpha_0^{i} - 3 L_0}\br{\beta^i}^2}, ~~~n = 0\\
\frac{\sigma_n^2}{2\br{\alpha_n^i - L_n}\br{\beta^i}^2}, ~~~n = 1, \ldots, N
\end{array}\right.
\end{equation}
and 
\begin{equation}\label{apdx5:24}
\eta_{15, n}^i := \left\{\begin{array}{l}
\frac{\epsilon + \alpha_0^{i}\beta^i}{2\beta^i}\norm{u_0^* - v_0^{i-1}}^2
 - \frac{\delta_0 + \alpha_0^{i}\beta^i}{2\beta^i}\norm{u_0^* - v_0^{i}}^2, ~~~n = 0\\
\frac{\alpha_n^i}{2T}\norm{u_n^* - v_{n,T}^{i-1}}^2
 - \frac{\delta_n + \alpha_n^i\beta^i}{2T\beta^i}\norm{u_n^* - v_{n,T}^i}^2, ~~~n = 1, \ldots, N.
\end{array}\right.
\end{equation}
\end{document}